\pdfoutput=1
\documentclass[11pt]{article}

\usepackage[final]{acl}

\usepackage{times}
\usepackage{latexsym}
\usepackage{float}
\usepackage[T1]{fontenc}
\usepackage[utf8]{inputenc}

\usepackage{microtype}

\usepackage{inconsolata}

\definecolor{green}{RGB}{0,200,0}

\usepackage{color}
\usepackage{placeins}

\usepackage{graphicx}

\usepackage{multirow}
\usepackage{algpseudocode}
\usepackage{stfloats}
\usepackage{float}
\usepackage{caption}
\usepackage{subfigure}
\usepackage{placeins}
\usepackage{subcaption}
\usepackage{natbib}
\usepackage{algorithm}
\usepackage{amsmath}  % For advanced math typesetting
\usepackage{amsfonts}
\usepackage{nicefrac}       % compact symbols for 1/2, etc.
\usepackage{hyperref} % For hyperlinks in the document
\usepackage{microtype}      % microtypography

\title{Optimize Weight Rounding via Signed Gradient Descent for the Quantization of LLMs}

\author{Wenhua Cheng \thanks{Correspondence:wenhua.cheng@intel.com} \and Weiwei Zhang \and   Haihao Shen \and  Yiyang Cai \\ {\bf Xin He} {\bf{\and}} {\bf Kaokao Lv}  {\bf{\and}}{\bf Yi Liu}   \\
Intel \\
}

\begin{document}
\maketitle
\begin{abstract}
Large Language Models (LLMs) have demonstrated exceptional proficiency in language-related tasks, but their deployment poses significant challenges due to substantial memory and storage requirements. Weight-only quantization has emerged as a promising solution, significantly reducing memory and storage needs without sacrificing too much performance.
In this study, we introduce SignRound, a method that leverages signed gradient descent (SignSGD) to optimize rounding values and weight clipping in just 200 steps. SignRound integrates the advantages of Quantization-Aware Training (QAT) and Post-Training Quantization (PTQ), delivering exceptional results across 2 to 4 bits while minimizing tuning costs and avoiding additional inference overhead. For example, SignRound achieved absolute average accuracy improvements ranging from 6.91\% to 33.22\% at 2 bits, as measured by the average zero-shot accuracy across 11 tasks. It also demonstrates strong generalization in recent models, achieving near-lossless 4-bit quantization in most scenarios.
% The source code will be made publicly available.
The source code is publicly available at \url{https://github.com/intel/auto-round}.
\end{abstract}

\section{Introduction}
\label{sec:Introduction}

In recent years, there has been a significant surge in the adoption of Large Language Models (LLMs), leading to their widespread deployment demand even on devices with constrained resources. However, deploying LLMs on these devices poses significant challenges due to their extensive memory and storage requirements. Additionally, the computational demands of these models create obstacles for real-time applications. Therefore, studying techniques such as quantization is crucial for enabling the efficient deployment of LLMs. Quantization techniques can be broadly categorized into two main types: quantization-aware training (QAT) \citep{esser2020learned, zhuang2021effective, lee2021cluster, liu2023llm} and post-training quantization (PTQ) \citep{nagel2019data, xiao2023smoothquant, frantar2022gptq, nagel2020up}.

QAT involves training the model with quantization in mind, using simulated lower-precision representations to allow the model to learn and adapt to the effects of quantization. This approach often results in better accuracy compared to PTQ. However, QAT has drawbacks, including increased training complexity, longer training times, and the need to tune hyperparameters. The application of QAT to LLMs can be particularly resource-intensive, despite recent efforts \citep{hu2021lora, dettmers2023qlora} to improve the efficiency of fine-tuning LLMs.

On the other hand, PTQ directly quantizes the model without any simulated training or fine-tuning. While PTQ is a more straightforward approach, it is susceptible to significant accuracy drops. This underscores the importance of further advancements in PTQ methods to enhance their accuracy preservation capabilities.

Quantization commonly applies to two types of tensors: activations and weights. Quantizing activations for LLMs can be challenging \citep{Wei_2023, xiao2023smoothquant, bondarenko2024quantizable}, making weight-only quantization a more practical option. Moreover, the main bottleneck in generating new tokens for LLMs often arises from memory bandwidth limitations \citep{kim2023squeezellm}, emphasizing the advantage of weight-only quantization. 

This study focuses on weight-only quantization. In quantizing weights, a critical step involves rounding, primarily achieved through rounding-to-nearest (RTN). RTN quantizes each weight independently by rounding it to the nearest integer, but it overlooks the relationships between weights and between weights and activations. Adaptive Rounding \citep{nagel2020up} explored the potential for an enhanced rounding strategy to improve accuracy. They approached the rounding task by formulating it as a quadratic unconstrained binary optimization problem and approximating the loss using a Taylor series expansion. However, relying solely on the second-order term may not yield accurate results, as rounding can significantly modify weights, making other order terms non-negligible.

\begin{figure*}[t]
\centering
\includegraphics[width=0.8\textwidth]{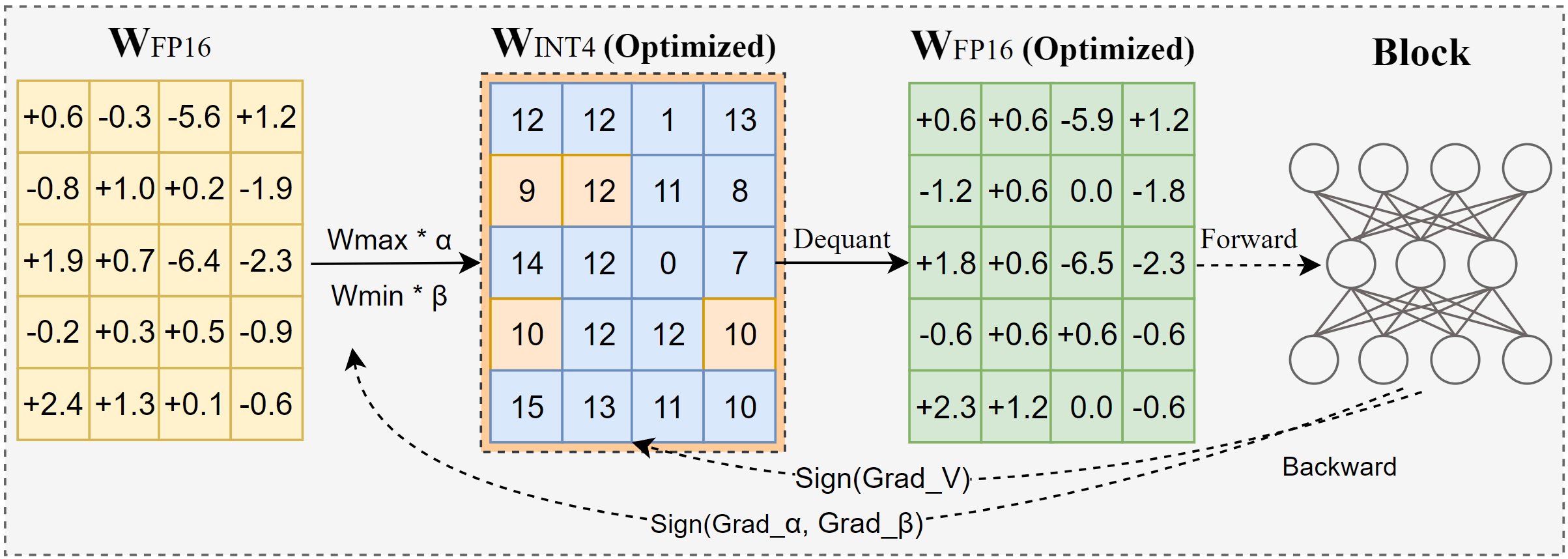}
\caption{An illustration of SignRound. Unlike the direct rounding in RTN, SignRound performs signed gradient descent to fine-tune the rounding and weight clipping through block-wise output reconstruction. After lightweight forward and backward steps, $\textbf{W}_{\text{INT4}}$ has been well optimized. Note that Quant and Dequant are two standard operations for quantization and dequantization respectively.}
\label{fig:signround_overview}
\end{figure*}

We select SignSGD\citep{balles2020geometry, li2023faster, safaryan2021stochastic} as our optimization method to approach the optimal rounding solution within a limited number of steps. The motivation behind this choice, which is elaborated in Section \ref{sec:Methodology}, stems from the well-defined boundaries of the solution space and the inherent simplicity of the method that necessitates only minimal hyperparameter tuning. Figure \ref{fig:signround_overview} provides an overview of our method. Our contributions primarily lie in three aspects:
\begin{itemize}

\item We introduce a concise yet effective method for optimizing the weight only quantization, combining the strengths of both QAT and PTQ. Our approach leverages SignSGD to tune the rounding with the weight clipping, without introducing any additional overhead during inference.

\item Our empirical results demonstrate a significant performance enhancement compared to recent works across various quantization configurations, ranging from 2-bit to 4-bit.

\item We demonstrate that SignRound's performance can be further enhanced by fine-tuning model-specific hyperparameters within a constrained space. Moreover, our method demonstrates strong generalization across various models and delivers nearly lossless results across the majority of scenarios using 4-bit quantization.
\end{itemize}

\section{Related Work}
\label{sec:RelatedWork}
\paragraph{Quantization Aware Training.}
 % QAT methods have gained widespread popularity in model compression, as they enable the fine-tuning process, often leading to superior accuracy compared to the PTQ method. In their work,~\citep{esser2020learned} proposed a novel approach that estimates and scales the task loss gradient at each weight and activation layer's quantizer step size, allowing for joint learning with other network parameters.~\citep{zhuang2021effective} put forward a progressive quantization scheme that involves quantizing activations after weights. Additionally, CPQ~\citep{lee2021cluster} effectively identified the optimal quantization grids while naturally encouraging the underlying full-precision weights to gather around those quantization grids cohesively during training. While QAT methods are popular in relatively small-scale models, their application in LLMs is limited due to the high computational cost associated with training or fine-tuning.
 
 QAT methods have gained widespread popularity in model compression, as they enable the fine-tuning process ~\citep{esser2020learned,zhuang2021effective,lee2021cluster}, often leading to superior accuracy compared to the PTQ method.

\paragraph{Post-training Quantization (PTQ).}
PTQ methods simplify the quantization process without the need for additional training.~\citep{nagel2019data,liu2021post,frantar2022optimal,hassibi1993optimal,yao2021hawq}. Given its low resource requirement, PTQ is particularly suitable for the quantization of Large Language Models.
% We will next focus on the quantization methods designed for LLMs, most of which fall under the category of PTQ.
% ,wang2019haq

\paragraph{Large Language Models Quantization.}
Significant strides have been made in addressing the pressing need for quantizing large language models (LLMs). GPT3.int8() \citep{dettmers2022gpt3} introduces a mixed-precision approach to preserve crucial channels in high precision.  
AQLM \citep{mao2024compressibility} builds upon Additive Quantization, a classic algorithm from the Multi-Codebook Quantization family, adapting it to LLM quantization. ZeroQuantV2 \citep{yao2024exploring} employs low-rank matrices to enhance model quality recovery. RPTQ \citep{yuan2023rptq} addresses range differences between channels by rearranging and quantizing them in clusters. LLM-QAT \citep{liu2023llm} employs QAT to enhance performance. Some other methods, such as SPIQ \citep{yvinec2023spiq}, SmoothQuant \citep{xiao2023smoothquant}, and Outlier Suppression+ \citep{Wei_2023}, utilize handcrafted equivalent transformations to mitigate quantization errors. These methods rely on the  model architecture to fuse the equivalent transformation operations. LRQ \citep{lee2024lrq} only needs to learn significantly fewer parameters while enabling the individual scaling of weights, thus boosting the generalization capability of quantized LLMs.

\paragraph{Weight Only Quantization.}
Weight-only quantization reduces the memory footprint and bandwidth demands by quantizing only the weights while retaining activations in floating-point precision, offering a promising balance between accuracy and compression. GPTQ~\citep{frantar2022gptq} optimizes weights using the Optimal Brain Surgeon technique \citep{hassibi1993optimal}, achieving low-bit quantization on LLMs with minimal tuning overhead. AWQ~\citep{lin2023awq} follows the equivalent transformation approach with additional tuning in a constrained space, sharing similar limitations with SmoothQuant \citep{xiao2023smoothquant}. TEQ~\citep{cheng2023teq} and OmniQuant~\citep{shao2023omniquant} both utilize a trainable equivalent transformation, while OmniQuant employs extra weight clip tuning. HQQ~\citep{badri2023hqq} accelerates quantization for large models by eliminating the need for calibration data, making the quantization process extremely fast. Some other works have incorporated optimization methods with extra inference overhead to improve quantization accuracy, such as dense-and-sparse decomposition techniques in SqueezeLLM~\citep{kim2023squeezellm} and EasyQuant~\citep{tang2023easyquant}, as well as nonuniform quantization methods in NUPES~\citep{yvinec2023nupes}, QuIP\#~\citep{tseng2024quip},\citep{gong2024makes}, AQLM~\citep{mao2024compressibility}, etc. Additionally, FineQuant~\citep{kim2023finequant} introduces a straightforward heuristic weight quantization approach that adaptively determines quantization granularity.
% AQLM\citep{mao2024compressibility} adapts the MAP-MRF optimization problem behind Additive Quantization to be instance-aware, considering layer calibration input and output activations to optimize weight quantization. 
In this work, we focus on approaches that do not introduce overhead during inference.
 
% QuIP\#\citep{tseng2024quip} introduces Hadamard incoherence and lattice codebooks to enhance the performance of ultra-low-bit weight quantization. Additionally, it employs fine-tuning to enhance fidelity to the original model. 
% Omniquant~\citep{shao2023omniquant} adopted trainable equivalent transformations. Although these approaches are effective, their applicability is limited due to the performance overhead involved during inference, because there is no chance to fuse the transformation scale to the model itself on certain model architectures.

\paragraph {Rounding Methods.}
Adaptive Rounding \citep{nagel2020up} has already showcased the potential of an advanced rounding strategy to enhance accuracy \citep{li2021brecq, wei2022qdrop}. They used the rounding task as a quadratic unconstrained binary optimization problem by approximating the task loss through a Taylor series expansion. However, considering only the second-order term may not yield accurate results. This is because the rounding value gets multiplied by a scaling coefficient during de-quantization, potentially introducing significant weight changes that make other order terms non-negligible. FlexRound \citep{lee2023flexround} introduces a more flexible approach to  rounding by incorporating element-wise division. However, it's not easily scalable to apply to LLMs due to the needs of specialized hyperparameters for each specific model and task. Furthermore, Oscillation-free \citep{liu2023oscillation} suggests that the introduction of learnable parameters might result in weight oscillation problems. AQuant \citep{li2022efficient} introduced a dynamic approach where the border becomes a function dependent on the activation value to reduce the quantization error of activation.

\paragraph {Signed Gradient Descent.} Signed gradient descent is not commonly utilized and is typically applied in specific scenarios, such as reducing communication costs. This is because signed gradient carries significantly less information compared to original gradient.  Recent studies have shed light on the advantages of sign-based methods over gradient descent in certain conditions. Balles et al. \citep{balles2020geometry} found that sign-based methods are preferable when the Hessian matrix is concentrated on its diagonal and the maximal eigenvalue is much larger than the average eigenvalue. Li et al. \citep{li2023faster} investigated a variant of sign-based gradient descent that exhibits faster convergence.  Safaryan et al. \citep{safaryan2021stochastic} proposed a stochastic sign descent with momentum, which converges under the standard bounded variance assumption with the optimal asymptotic rate. These findings contribute to a better understanding of the potential benefits and applications of signed gradient descent methods.

\section{Methodology}
\label{sec:Methodology}

We begin with an overview of quantization before delving into the specifics of our approach. The following operations can be utilized to quantize and dequantize the weights $\textbf{W}$:
\begin{equation}
\mathbf{\widetilde{W}} = s*clip(\left\lfloor\frac{\mathbf{W}}{s}+zp \right\rceil,n,m),n,m \in \mathbb{N}
\label{eq 1}
\end{equation}
where the rounding operation $\left \lfloor \cdot \right \rceil $
 is typically performed using the RTN method. Although RTN is a straightforward approach, it quantizes each element independently, which results in the loss of the ability to model the correlation among different weights or activations. The $s$ represents the quantization scale, which can be obtained using the following equation, and $zp$ is the zero point.

\begin{equation}
\label{eq 2}
s = \frac{max(\mathbf{W})-min(\mathbf{W})}{2^{bit}-1}
\end{equation}

\begin{algorithm}[t]
\caption{SignRound}
\label{algorithm signround}
\textbf{Input:} Calibration Data $\mathcal{D}$, learning rate $lr$, total steps $T$, Model $M$, block module $m_w$ with weights $w$, batch size $bs$ 

\textbf{Output:}  $best\_V$, $best\_\alpha$, $best\_\beta$
\begin{algorithmic}[1]
\State  $V \gets 0$, $\alpha \gets  1.0$,  $\beta \gets  1.0$ , $best\_l \gets maximum$
\For{$i \gets 0$ to $T$}
        \State $d \gets \textit{draw $bs$ samples}$
        \State $x \gets M(d)_m $  \Comment{get the inputs of m} 
        \State $y_f \gets  m_w(x) $  \Comment{get the output of original module} 
        \State $\widetilde{w}\gets qdq(w,\alpha, \beta, V)$  \Comment{quantize and dequantize w via Eq.\ref{eq 3}}
        \State $y_q \gets  m_{\widetilde{w}}(x)$ \Comment{get the output of quantized module}
        \State $loss \gets mse(y_q, y_f) $ \Comment{get the loss via Eq.\ref{eq 5}}
        \State $loss.backward()$
        \If{$loss< best\_l$}
            \State $best\_V, best\_\alpha, best\_\beta \gets V, \alpha, \beta$
            \State $best\_l \gets loss $
       \EndIf  
        \State \textit{update $\alpha$, $\beta$ and V via SignSGD optimizer }

\EndFor
\end{algorithmic}
\end{algorithm}

In order to improve the efficacy of the rounding quantization operation, we build upon prior research \citep{nagel2020up} by introducing a trainable parameter $V$ to adjust the rounding values.

\begin{equation}
\widetilde{\mathbf{W}} = s*clip(\left \lfloor\frac{\mathbf{W}}{s}+zp +\mathbf{V} \right \rceil,n,m),n,m \in \mathbb{N}
\label{eq 3}
\end{equation}

Additionally, following recent works \citep{lin2023awq,shao2023omniquant}, we introduce two additional trainable parameters, denoted as $\alpha  \in [0,1 ]$ and $\beta 
 \in [0,1 ]$, to fine-tune the scale of weight clipping. These parameters are incorporated into the equations as follows:
\begin{equation}
\label{eq 4}
s = \frac{max(\mathbf{W})*\alpha-min(\mathbf{W})*\beta}{2^{bit}-1}
\end{equation}
These modifications enable a more adaptable quantization process. We utilize block-wise output reconstruction to train these parameters via optimizer, thus framing the optimization as follows.

\begin{equation}
\label{eq 5}
\min_{\alpha, \beta, \mathbf{V}} \|\mathbf{WX}-\mathbf{\widetilde{W}X}\|_F^2 
\end{equation}
where $\textbf{X}$ is the input of the block and $||\cdot||_F$  denotes the Frobenius norm.

Our method distinguishes itself primarily by leveraging SignSGD, which optimizes parameters based on the sign of the gradients as follows:

\begin{equation}
\label{eq 6}
\mathbf{W}_{t+1} = \mathbf{W}_t - lr_t * sign(\mathbf{g}_t)
\end{equation}
where t represents the step, lr is the learning rate and g denotes the gradient.
 The motivation is detailed below.  \textit{\textbf{Firstly}}, the optimal values for up and down rounding typically reside in a large region rather than a single float, as only the threshold for altering the rounding value is significant. This eliminates the necessity for the gradient magnitude to converge precisely to a single point. \textit{\textbf{Secondly}}, due to the confined boundaries, i.e.$[-0.5, 0.5]$ for rounding and $[0, 1]$ for weight clipping, SignSGD allows efficient navigation of this space within a limited number of steps. In contrast, optimizers like Adam \citep{kingma2014adam} may struggle due to significant variations in gradient magnitude, making it challenging to converge to the optimal value within a restricted number of steps. \textit{\textbf{Thirdly}}, SignSGD is inherently intuitive, facilitating easy adjustment of the step size (learning rate). For example, we employed the same optimizer hyperparameters across all experiments unless explicitly stated, consisting of 200 steps and a learning rate of 5e-3 with linear weight decay. Based on Eq. \ref{eq 6}, the maximum adjustment for each parameter is the sum of the learning rates over all steps, that is, $200 \times 0.005 / 2 = 0.5$. As a result,the adjustment can cover a range of [-0.5,0.5] when initialized at 0 for rounding, and a range of [0.5,1.0] when initialized at 1 and clipped to $\leq 1.0$ for weight clipping, which works well in practice. \textit{\textbf{Fourth}}, SignSGD distinguishes itself by its lightweight design, demanding fewer memory and computational resources than  optimizers like Adam\citep{kingma2014adam}.

 Figure \ref{fig:signround_overview} provides an illustration of our approach. And the Pseudocode \ref{algorithm signround} presents more details of SignRound.
 
\begin{table*}
\centering
\scalebox{0.82}{
\setlength{\tabcolsep}{2pt}{
    \begin{tabular}{|cl|cccc|}
    \hline
        \textbf{Config} & \textbf{Method} & \textbf{Mistral-7B} & \textbf{V2-7B} & \textbf{V2-13B} & \textbf{V2-70B} \\ \hline
         & 16 bits & 63.30 & 57.98 & 61.42 & 66.12 \\ 
        \hline
         \multirow{7}*{\fontsize{9}{12}\selectfont W4G-1} & RTN & 58.84 & 55.49 & 60.46 & 65.22 \\
         ~ & GPTQ & 61.37 & 56.76 & 59.79 & 65.75 \\
        ~ & AWQ & 61.36 & 57.25 & 60.58 & \textbf{66.28} \\
         ~ & HQQ & 58.40 & 46.05 & 46.82 & 57.47 \\
        ~ & Omni & 60.52 & 56.62 & 60.31 & 65.80 \\
        ~ & Ours & \textbf{62.33} & \textbf{57.48} & \textbf{61.20} & 66.27 \\
        \cline{3-6}
        ~ & Ours*   & \textbf{62.64} & \textbf{57.52} & \textbf{61.23} & 66.27 \\
        \hline
         \multirow{7}*{\fontsize{9}{12}\selectfont W4G128} & RTN & 62.36 & 56.92 & 60.65 & 65.87 \\
         ~ & GPTQ & 62.32 & 56.85 & \textbf{61.00} & 66.22 \\
        ~ & AWQ & 62.16 & 57.35 & 60.91 & 66.23 \\
         ~ & HQQ & \textbf{62.75} & 57.41 & 60.65 & 66.06 \\
        ~ & Omni & 62.18 & 57.30 & 60.51 & 66.02 \\
        ~ & Ours & 62.62 & \textbf{57.57} & 60.85 & \textbf{66.39} \\
        \cline{3-6}
        ~ & Ours* & \textbf{62.87} & \textbf{57.97} & 60.90 & \textbf{66.41} \\
        \hline
    \end{tabular}

    \begin{tabular}{|cl|cccc|}
     \hline
        \textbf{Config} & \textbf{Method} & \textbf{Mistral-7B} & \textbf{V2-7B} & \textbf{V2-13B} & \textbf{V2-70B} \\ \hline
         & 16 bits & 63.30 & 57.98 & 61.42 & 66.12 \\
         % \cline{2-6}
         \hline
        \multirow{7}*{\fontsize{10}{12}\selectfont W3G128} & RTN & 58.20 & 53.81 & 58.57 & 64.08 \\
        ~ & GPTQ & 59.91 & 54.14 & \textbf{59.58} & 65.08 \\
        ~ & AWQ & 59.96 & 55.21 & 58.86 & 65.12 \\
        ~ & HQQ & 59.33 & 54.31 & 58.10 & 64.80 \\
        ~ & Omni & 58.53 & 54.72 & 59.18 & 65.12 \\
        ~ & Ours & \textbf{60.43} & \textbf{56.68} & 59.44 & \textbf{65.31} \\
        \cline{3-6}
        ~ & Ours* & \textbf{60.96} & \textbf{56.68} & \textbf{59.78} & \textbf{65.59} \\
        \hline
        \multirow{7}*{\fontsize{10}{12}\selectfont W2G128} & RTN & 30.52 & 29.94 & 33.51 & 38.14 \\
        ~ & GPTQ & 39.61 & 35.37 & 42.46 & 28.47 \\
        ~ & AWQ & 30.06 & 30.10 & 32.16 & 32.23 \\
         ~ & HQQ & 31.41 & 29.87 & 35.28 & 37.42 \\
        ~ & Omni & 32.17 & 40.74 & 46.55 & 51.31 \\
        ~ & Ours & \textbf{52.71} & \textbf{48.64} & \textbf{53.46} & \textbf{61.69} \\
        \cline{3-6}
         ~ & Ours* & \textbf{53.01} & \textbf{50.34} & \textbf{54.16} & \textbf{61.77} \\
        \hline
    \end{tabular}
}}
\caption{\label{tab:llama and mistral at W2-W4}
Average accuracies ($\uparrow$) across 11 tasks, as detailed in Section \ref{sec:tasks}, for LLaMA and Mistral models at W2-W4. 'Our*' denotes the highest accuracy achieved among the 8 hyperparameter choices, outlined in Section \ref{sec:our* detail}, whereas for the 70B model, we tested only a few options.}

\end{table*}

\section{Experiments}

This section presents a comprehensive evaluation of SignRound from multiple perspectives. We begin with a brief overview of the LLM architectures and tasks included in our assessment. Next, we provide a detailed comparison between our method and several existing approaches, emphasizing the unique advantages of SignRound. Furthermore, we conduct ablation studies to reinforce the efficacy of our choices and investigate the sensitivity of hyperparameters. Lastly, we evaluate the generation ability of our method across various recent models. The tuning cost comparisons are provided in Appendix \ref{sec:runtime}.

\subsection{Experimental Settings}
\label{sec:4.1}

\begin{table*}[t]
\centering
\scalebox{0.86}{
\setlength{\tabcolsep}{1pt}{
\begin{tabular}{ll|cccccccccccc}
\hline
  \textbf{Config} &  \textbf{Method} &  \textbf{Mmlu} &  \textbf{Lamb.}  &  \textbf{Hella.} &  \textbf{Wino.} &  \textbf{Piqa}  &  \textbf{Truth.} &  \textbf{Open.}  &  \textbf{Boolq}  &  \textbf{RTE} &  \textbf{ARC-e}  &  \textbf{ARC-c.}  &  \textbf{Avg.} \\ \hline
    \multirow{8}*{W4G-1} & 16 bits  & 61.35 & 75.68 & 61.27 & 74.03 & 80.79 & 28.03 & 32.80 & 83.67 & 67.51 & 80.81 & 50.34 & 63.30 
    \\
    
    ~ & RTN  & 55.92 & 66.10 & 59.01 & 71.35 & 80.14 & 24.85 & 29.00 & 79.17 & 57.76 & 77.95 & 45.99 & 58.84 
    \\ 
    
    ~ & GPTQ & 58.22 & 73.45 & 59.47 & \textbf{74.03}  & \textbf{80.20}  & 26.93 & 31.00 & 81.50 & 64.98 & 78.24 & 47.01 & 61.37 
      \\ 
       
    ~ & AWQ  & 57.20 & 71.45 & 59.21 & 73.64 & 79.43 & 25.34 & 30.40 & \textbf{82.69}  &\textbf{68.95}  & 79.25 & 47.44 & 61.36 
      \\
    
    ~ &  HQQ & 52.65  & 66.58  & 59.09  & 70.56  & 79.60  & 23.13  & 27.80  & 80.03  & 59.57  & 77.02  & 46.33  & 58.40  \\
    
    ~ &  Omni & 57.52  & 70.00  & 60.27  & 72.93  & 79.87  & 23.99  & 30.80  & 81.53  & 63.90  & 78.54  & 46.42  & 60.52  \\
       
    ~ &  Ours   & \textbf{59.52}  & \textbf{73.76}  & \textbf{60.75}  & 73.32 & 80.09 & \textbf{27.17}  & \textbf{33.00}  & 82.02 & 66.07 & \textbf{80.47}  & \textbf{49.49}  & \textbf{62.33}  \\ 
    
    \cline{3-14}
    
    ~ &  Ours* & \textbf{60.00} & 73.30 & 60.57 & \textbf{74.35} & 80.09 & \textbf{27.91} & 32.20 & \textbf{83.52} & 67.51 & 79.92 & \textbf{49.66} & \textbf{62.64}   \\
    
    \hline
    
    & 16 bits & 61.35 & 75.68 & 61.27 & 74.03 & 80.79 & 28.03 & 32.80 & 83.67 & 67.51 & 80.81 & 50.34 & 63.30 \\
        
    & RTN  & 23.45 & 0.14 & 27.43 & 49.64 & 54.30 & 24.24 & 15.20 & 38.69 & 51.99 & 29.08 & 21.59 & 30.52 \\ 
    
    & GPTQ & 25.23 & 30.47 & 38.28 & 53.83 & 64.91 & 24.11 & 17.40 & 58.29 & 50.90 & 47.77 & 24.57 & 39.61 \\ 
       
     W2G128 & AWQ  & 25.38 & 0.00 & 25.71 & 52.01 & 51.58 & 23.99 & 17.60 & 37.83 & 47.29 & 26.98 & 22.27 & 30.06 \\
    
     & HQQ & 23.35 & 0.85 & 27.77 & 51.62 & 56.69 & \textbf{26.68} & 15.80 & 40.55 & 53.43 & 28.62 & 20.14 & 31.41 \\
     
     & Omni & 23.24 & 5.38 & 29.38 & 49.72 & 56.09 & 26.32 & 16.60 & 41.99 & 52.71 & 32.11 & 20.39 & 32.17 \\
       
     &  Ours  & \textbf{40.46} & \textbf{58.61} & \textbf{50.87} & \textbf{62.90} & \textbf{75.84} & 24.85 & \textbf{22.80} & \textbf{78.56} & \textbf{57.04} & \textbf{70.88} & \textbf{37.03} & \textbf{52.71 } \\

     \cline{3-14}
     
     &  Ours* & \textbf{43.72} & \textbf{59.75}  & \textbf{51.87} & \textbf{64.25} & 75.14 & 24.72 & \textbf{23.60} & 75.78 & 55.23 & \textbf{71.80} & \textbf{37.20} & \textbf{53.01} \\

\hline
\end{tabular}}
}
\caption{\label{tab:mistral at W4G-1 and W2G128}
Detailed accuracies($\uparrow$) across 11 tasks(0-shot) of Mistral models at W4G-1 and W2G128.  'Our*' denotes the highest accuracy achieved among the 8 hyperparameter choices, outlined in Section \ref{sec:our* detail}. Appendix \ref{sec:Appendix other results} provides more detailed data.}
\end{table*}

\begin{table}
\centering
\scalebox{0.90}{
\setlength{\tabcolsep}{2pt}{
\begin{tabular}{c|ll|ccc}
\hline
    \textbf{Model} & \textbf{Method} & \textbf{Steps} & \textbf{Mistral-7B} & \textbf{V2-7B} & \textbf{V2-13B} \\
  \hline 
  \multirow{7}*{\fontsize{10}{12}\selectfont W4G-1} & \multirow{3}*{Flex} & 200 & 58.93 & 56.10 & 60.06 \\
    ~ & ~ & 1000 & 60.62 & 56.98 & 60.29 \\
    ~ & ~ & 5000 & 60.94 & 57.49 & 60.69 \\
    \cline{2-3}
    ~ & \multirow{2}*{Ada}  &  200  & 58.30 & 55.06 &  59.86  \\
    ~ & ~  & 1000 & 58.38  & 55.05 & 59.92  \\
    \cline{2-3}
    ~ & \multirow{2}*{Ours} & 200 & 62.33 & 57.48 & 61.20  \\
    ~ & ~ & 200* & \textbf{62.64} & \textbf{57.52} & \textbf{61.23} \\
    % ~ &  & 1000 \\
  \hline
  \multirow{6}*{\fontsize{10}{12}\selectfont W2G128} & \multirow{2}*{Flex} & 200 & 30.10 & 30.01 & 30.66 \\ 
    ~ & ~ & 1000 & 30.16 & 31.26 & 32.29 \\
    \cline{2-3}
    ~ & \multirow{2}*{Ada}  &  200 & 30.74 & 30.21 & 30.36  \\
    ~ & ~  & 1000 & 30.84 & 30.30 & 30.02 \\
    \cline{2-3}
    ~ & \multirow{2}*{Ours} & 200 & 52.71 & 48.64 & 53.46 \\
    ~ & ~ & 200* & \textbf{53.01} & \textbf{50.34} & \textbf{54.16} \\
    % ~ & Ours* & - \\
  \hline
\end{tabular}}}
\caption{\label{tab:Rounding_comparing}
Comparing with some other rounding methods, the average accuracies ($\uparrow$) across 11 tasks (detailed in Section \ref{sec:tasks}) for Mistral and LLaMA models at W4G-1 and W2G128.}
\end{table}

\paragraph{Evaluation and Tasks.}
\label{sec:tasks}
We evaluate multiple language tasks to address the task-agnostic setting. Specifically, we present the average accuracy results for 11 zero-shot tasks, including HellaSwag \citep{Zellers_2019}, WinoGrande \citep{sakaguchi2021winogrande}, PIQA \citep{bisk2020piqa}, LAMBADA \citep{paperno2016lambada}, TruthfulQA \citep{lin2022truthfulqa}, OpenBookQA \citep{mihaylov2018can}, BoolQ \citep{clark2019boolq}, RTE \citep{dagan2010recognizing}, ARC-Easy, ARC-Challenge \citep{clark2018think}, and MMLU \citep{hendrycks2020measuring}. We use lm-eval-harness \citep{eval-harness} for all the above tasks. Furthermore, we complement our evaluation with perplexity (PPL) analysis on Wikitext2 \citep{merity2016pointer}, PTB \citep{marcus1993building}, and C4 \citep{raffel2020exploring}, following the implementation\footnote{https://github.com/IST-DASLab/gptq} of GPTQ and Wikitext2 \citep{merity2016pointer} using lm-eval-harness \citep{eval-harness}. However, we argue that perplexity is notably influenced by outliers, as illustrated in Table \ref{tab:llama mistral ppl} for different algorithms. This susceptibility likely arises from the mathematical expression
$\mathrm{PPL}(X)=\exp\left(-\frac{1}{t}\sum_{i=1}^{t}\log p_\theta(x_i|x_{<i})\right)$, where assigning a low probability to even one token can significantly inflate the perplexity score. Consequently, we prioritize the accuracy of the 11 tasks mentioned above as the primary metric, with perplexity data serving as supplementary reference.

\paragraph{Quantization Configurations.}
In alignment with GPTQ \citep{frantar2022gptq}, our focus is specifically on weight-only quantization, targeting the linear layers within transformer blocks. Layers such as the embedding layer and typically the last linear layer like 'lm-head' are excluded from the quantization process. 
Our evaluation primarily centers on W4G-1, W4G128, W3G128 and W2G128 configurations, where W4 indicates quantizing weights with 4 bits and G represents finer-grained grouping as described in \citep{park2022nuqmm, frantar2022gptq}. We adopt asymmetric quantization.
To mitigate overfitting on the WikiText and C4 datasets, for all the methods that need calibration, we randomly select 512 calibration samples with the same seed from the readily available pile-10k dataset \footnote{\url{https://huggingface.co/datasets/NeelNanda/pile-10k}} , which comprises the first 10k samples from pile \citep{gao2020pile}. We used a sequence length of 2048 for calibration, while for other methods, we adhere to their official settings.

\paragraph{Large Language Models.}
We compare different algorithms on commonly used models such as LLaMA-V1 \citep{touvron2023llama}, LLaMA-V2 \citep{touvron2023llamav2}, and Mistral-7B-v0.1 \citep{jiang2023mistral}. Our comparison covers a wide range of LLM parameters, ranging from 7B to 70B, to ensure comprehensive coverage and analysis.

\paragraph{SignRound Hyperparameters.}
Unless explicitly stated, the tuning process involved adjusting each block for 200 steps with a learning rate of $5 \times 10^{-3}$, a batch size of 8, and linear learning rate decay. Additionally, we employed automatic mixed precision (AMP) to accelerate the tuning.

\subsection{Comparing With Recent Methods}
\label{sec:4.2}

In this section, we compare our methods with those that have already demonstrated remarkable results and impose no additional overhead on our tested models in weight-only quantization for LLMs, including GPTQ \citep{frantar2022gptq}, AWQ \citep{lin2023awq}, HQQ \citep{badri2023hqq}, OmniQuant \citep{shao2023omniquant} with a naive method RTN.

To ensure fair comparison as much as possible, we enabled act-order and true-sequential in GPTQ and also activated static\_group in scenarios with   group\_size. The notation GPTQ$^+$ indicates that we adjusted the random seed or data pre-processing to address issues related to the non-positive definite Hessian matrix or other issues. For OmniQuant\citep{shao2023omniquant}, we adhere to the official settings, which include running for 20 epochs including W2G128 for saving time and disabling 'let'. We conducted calibration tests using sample sizes of 512 and 128, as well as a sample size of 512 with a batch size of 4. Our findings show that using a sample size of 512 typically results in comparable or slightly higher performance for models less than or equal to 13B. Therefore, we present the results based on the sample size of 512.  For 70B models, due the the Not a Number (NAN) loss issue and to reduce the tuning cost of OmniQuant, we adopted 128 samples for calibration.

\begin{table*}
\centering
\scalebox{0.9}{
\setlength{\tabcolsep}{2pt}{
\begin{tabular}{ll|cccccccccc}
\hline
\textbf{Config}  &  \textbf{Model} & \textbf{2.5e-3} & \textbf{5e-3}  & \textbf{7.5e-3} & \textbf{1e-2}  & \textbf{1.25e-2} & \textbf{1.5e-2} & \textbf{1.75e-2} & \textbf{2e-2} & \textbf{SignSGD} \\
  \hline 
\multirow{3}*{\fontsize{10}{12}\selectfont W4G-1} & Mistral-7B  & 61.82 & 61.16   & 61.30   & 60.69   &  60.80  & 61.07   &    61.53 &  61.23  &  \textbf{62.33}  \\

~ & V2-7B  & 56.79  &  57.45 &  57.09  &  57.28  &  56.88  &  57.24  &  57.40  &  57.10  &  \textbf{57.48}  \\

~ & V2-13B  & 60.58 &  60.73 & 60.76  &  60.86 &  61.02  &  60.79   &  61.06  &  60.85  &  \textbf{61.20}  \\  
\hline

\multirow{3}*{\fontsize{10}{12}\selectfont W2G128} & Mistral-7B  & 37.12 &  40.37 & 41.11  & 42.02 &  42.86  &  43.55  &  43.44  &  42.44  &  \textbf{52.71}  \\

~ & V2-7B  & 42.26  & 44.64  &  45.08  & 45.04   &  45.15  &  43.13  & 38.71  &  35.73  &  \textbf{48.64}  \\

~ & V2-13B  & 47.81  & 50.01  & 49.55   &  50.80  & 48.67   & 51.94   & 38.28  &  34.67 &  \textbf{53.46}  \\ \hline
\end{tabular}}}
\caption{\label{tab:signgd_adam}
Comparison of Adam optimizer with various learning rates against the SignSGD optimizer.. The average accuracies($\uparrow$) across 11 tasks (detailed in Section \ref{sec:tasks}) for Mistral and LLaMA models at W4G-1 and W2G128.}
\end{table*}

\begin{table*}
\centering
\scalebox{0.9}{
\setlength{\tabcolsep}{2pt}{
\begin{tabular}{l|ccc|ccc}
\hline
\textbf{Config} & \textbf{Mistral-7B} & \textbf{V2-7B} & \textbf{V2-13B} & \textbf{Mistral-7B} & \textbf{V2-7B} & \textbf{V2-13B} \\ \hline
 & \multicolumn{3}{c|}{W4G-1}& \multicolumn{3}{c}{W2G128} \\ 
\hline
RTN & 58.84& 55.49  &60.46 &30.52 & 29.94 & 33.51 \\

Weight clip only &61.10 & \textbf{57.41}& 60.10& 46.60 & 40.53& 49.77 \\

Rounding only &\textbf{61.62} & 56.74 &\textbf{60.64} &\textbf{52.32} &\textbf{49.14} &\textbf{54.41} \\
\hline
Default  &\textbf{62.33} &\textbf{57.48} &\textbf{61.20} &\textbf{52.71} &48.64 & 53.46\\
\hline
\end{tabular}}}
\caption{\label{tab:round_weight_clip_tuning}
Ablation study of round tuning and weight clip tuning. The average accuracies($\uparrow$) across 11 tasks(detailed in Section \ref{sec:tasks}) for Mistral and LLaMA models at W4G-1 and W2G128.}
\end{table*}

 We present the summary results of Mistral-7B and LLaMAV2 in Table \ref{tab:llama and mistral at W2-W4}, detailed results of Mistral-7B in Table \ref{tab:mistral at W4G-1 and W2G128}, and additional detailed results are provided in Appendix \ref{sec:Appendix other results} due to space constraints. In summary, our approach demonstrated superior performance compared to GPTQ \citep{frantar2022gptq}, achieving scores of 30/32, AWQ \citep{lin2023awq} with 27/32, HQQ \citep{badri2023hqq} with 15/16, and OmniQuant \citep{shao2023omniquant} with a score of 29/32 across LLaMAV1/LLaMAV2/Mistral-7B on various quantization settings, including W4G-1, W4G128, W3G128, and W2G128. These evaluations were based on the average accuracies of 11 zero-shot tasks.

It's worth noting that as the bit depth decreases, the advantages of SignRound become more notable.  For example, as shown in Table \ref{tab:mistral at W4G-1 and W2G128}, SignRound could yield absolute average accuracy improvements ranging from 6.91\% to 33.22\% at W2G128.

\label{sec:our* detail} Moreover, we can enhance the performance by tuning the model's hyperparameters from a selection of eight choices, denoted as ours*. These choices include steps (200, 1000), weight clip learning rate (1.0/steps, 2.0/steps), and the option to either enable or disable quantized inputs, which refers to utilizing the output from the previous quantized block or the previous original block.

% Typically, the loss incurred from 2-3-bit ultra-low bit quantization is considered relatively unacceptable. Through experimental analysis of 4-bit quantization, it has been observed that ungrouped quantization would result in even more severe accuracy loss. Consequently, when quantifying at 2-3 bits, we solely refer to the results obtained from Grouping 128.

% Table \ref{llama and mistral at W2-W4} showcases the average accuracy data across 2-4 bits for different models. SignRound demonstrates advantages in 12 out of 16 scenarios. In the 2-bit quantization scenario, SignRound consistently exhibits higher accuracy across all tasks and models. Conversely, other algorithms generally exhibit lower accuracy across the 11 tasks and perform poorly on some datasets (e.g., lamabada\_openai), making them unsuitable for ultra-low-bit quantization scenarios. This underscores the significant potential of applying signed gradient descent for fine-tuning in ultra-low-bit quantization, leveraging a rounding strategy. Extensive experimental data also reveals that various algorithms perform better and achieve higher accuracy at grouping 128. This confirms that fine-grained grouping in weight quantization can effectively mitigate the impact of outliers on quantization error. Under the constraint of fixed learning parameters, all SignRound quantized W4G128 models, except mistral, meet the criterion of quantization accuracy loss within 1\%.

\subsection{Comparing with Rounding Methods}
In this section, we conduct a comparative analysis between SignRound, FlexRound\citep{lee2023flexround}, and AdaRound\citep{nagel2020up}. Notably, during the experiment, there is no formal official implementation available for FlexRound and AdaRound for LLMs. Hence, we reference the implementations \footnote{\url{https://openreview.net/forum?id=-tYCaP0phY_}} \footnote{\url{https://github.com/quic/aimet}} for further details.  However, it's important to highlight that due to the lack of AMP support and other optimizations, the implementation is notably slow, especially when adhering to the official settings, which involve tuning 5000 steps, as presented in Table \ref{tab:rounding_runtime}. Therefore, our comparison is limited to models of size 13B or smaller. We set the learning rate to 2e-4 for LLaMA-v2-7b and Mistral-7B, and 1e-4 for LLaMA-v2-13b to align with the official settings as closely as possible. As shown in Table \ref{tab:Rounding_comparing}, SignRound achieves better results in just 200 steps compared to the 5000 steps required by other rounding methods.

\subsection{Ablation Studies}

\paragraph{SignSGD versus Adam.} To validate the effectiveness of SignSGD, Table \ref{tab:signgd_adam} compares it with the Adam optimizer \cite{kingma2014adam}. SignSGD employs a fixed learning rate of 5e-3 throughout all experiments, comprising 200 steps, with linear weight decay. For Adam, we explored learning rates ranging from 2.5e-3 to 2e-2. We choose to quantize models of 13B or less with W4G-1 due to the experiment's cost. SignSGD demonstrated a distinct advantage in average accuracy metrics across 11 tasks, which demonstrate the unique advantage of signed gradient descent in this scenario.

\paragraph{Round and Weigh Clip Tuning.} To validate the contributions of rounding tuning and weight clip tuning, we conducted ablation studies on three models with two quantization configurations. As shown in Table \ref{tab:round_weight_clip_tuning}, each component provides benefits over RTN, with rounding tuning offering greater advantages. However, when combined, weight clip tuning can sometimes result in lower accuracy in certain  cases at W2G128.

\begin{table*}
\centering
\scalebox{0.9}{
\begin{tabular}{l|cccccccc}
\hline
\textbf{Model}  & \textbf{Seqlen\_512} & \textbf{Samples\_128}  & \textbf{Batch\_4} & \textbf{Steps\_100} & \textbf{Steps\_1000}   & \textbf{LR\_1e-2} & \textbf{Default} \\
  \hline 
Mistral-7B &  60.32  &  61.82  &  61.78  &   61.06  &  \textbf{62.58}  &  61.27   &  62.33  \\

V2-7B      &  \textbf{57.91}  &  56.41  &  57.21   &  57.10  &  57.19  &  55.89   &  57.48  \\

V2-13B     &  60.88  &  60.87  &  \textbf{61.21}   &  60.80  &  61.01  &   61.03  &  61.20  \\
\hline
\end{tabular}}
\caption{\label{tab:hyperparameter sensitivity}
Ablation study of hyperparameter sensitivity. The average accuracies($\uparrow$) across 11 tasks(detailed in Section \ref{sec:tasks}) for LLaMA models at W4G-1.}
\end{table*}

\paragraph{Hyperparameters Sensitivity.} To validate the sensitivity of hyperparameters in SignRound, we conducted ablation studies on sequence length for calibration, the number of samples for calibration, tuning batch size, tuning steps, and tuning learning rate. The results are presented in Table \ref{tab:hyperparameter sensitivity}. Overall, our default hyperparameters achieved balanced results.

% this is because usually the optimal value for up and down rounding acts over a range that is a larger region and is not limited to a single point, implying that there is no need for the gradient magnitude to converge to a precise value, and that focusing on the tendency to converge will be sufficient to satisfy the need for rounding.

\begin{table*}
\centering
\scalebox{0.9}{
\setlength{\tabcolsep}{2pt}{
\begin{tabular}{ll|cc}
\hline
    \textbf{Model} & \textbf{Method} &  \textbf{Average Acc} & \textbf{Variation \%} \\ \hline
        \multirow{2}*{Gemma-2b} & BF16 & 53.69  & - \\
        ~ & Ours & 53.40 & -0.54\% \\ \hline
        \multirow{2}*{Llama-2-7b-chat-hf} & FP16  & 59.01 & - \\
        ~ & Ours  & 58.97 & -0.07\% \\ \hline
        \multirow{2}*{Llama-3-8B-Instruct}
        & BF16  & 63.52  & - \\
        ~ & Ours   & 63.12  &  -0.63\% \\ \hline
        \multirow{2}*{Mistral-7B-Instruct-v0.2} & BF16  & 66.47 & - \\
        ~ & Ours  & 66.21 & -0.39\% \\ \hline
        \multirow{2}*{Mixtral-8x7B} & BF16  & 66.98 & - \\
        ~ & Ours  & 66.33 & -0.97\% \\ \hline
        \multirow{2}*{Mixtral-8x7B-Instruct} & BF16  & 70.00 & - \\
        ~ & Ours & 69.77 & -0.33\% \\ \hline
        \multirow{2}*{Phi-3-mini-4k-instruct} & BF16  & 66.62 & - \\
        ~ & Ours  & 66.33 & -0.44\% \\ \hline
\end{tabular}}
}
\caption{\label{tab:generalization of LLMs}
 The average accuracies($\uparrow$) across 11 tasks(detailed in Section \ref{sec:tasks})) with 1000 steps for LLMs at W4G128. Table \ref{tab:details of LLM generalization} provides the detailed data.}
\end{table*}

\subsection{Generalization to Other Models}
\label{sec4.3}
% With only 200 tuning steps, SignRound has already achieved satisfactory quantization accuracy at W4. Furthermore, as the number of tuning steps increased, the quantization performance of SingRound improved even further. 

To assess the generalization of our method on LLMs, we evaluate SignRound on various mainstream LLMs such as Gemma \citep{team2024gemma}, Phi \citep{li2023textbooks}, Mistral \citep{jiang2023mistral}, Mixtral \citep{jiang2024mixtral} and Llama3 \citep{meta2024llama3}. Table \ref{tab:generalization of LLMs} demonstrated that all int4 models maintained an accuracy drop within 1\% of FP16 or BF16 accuracy by employing 1000 tuning steps and model wise hyperparameters among 4 choices detailed in Section \ref{sec:4.1}. The detailed results are provided in Appendix \ref{sec:Appendix other results}. Notably, the generalization experiments utilized an updated version (0.4.0+) of lm-eval-harness \citep{eval-harness} and real quantized models, which may result in minor discrepancies compared to other benchmark data.

\section{Conclusions}

In this paper, we introduce SignRound, an efficient and concise approach for optimizing weight rounding in the quantization of large language models. SignRound employs signed gradient descent for tuning rounding value and weight clipping in 200 steps, completing the quantization of LLAMA-V2-70B in approximately 2.5 hours. Our extensive experiments show that SignRound outperforms other quantization methods across various models and weight bits in the majority of scenarios. Additionally, SignRound shows promising generation capabilities in recent models and achieves enhanced performance through model-specific hyperparameter tuning.

\newpage
% Moreover, as the input sequence size increases, the KV cache emerges as the primary bottleneck. In this context, weight-only quantization proves inadequate as a compression scheme. Thus, the combination of SignRound with KV cache pruning represents a promising avenue for further investigation.

% In the future, we intend to expand SignRound to encompass non-uniform quantization and explore mixed-precision quantization. We expect that this expansion will lead to further improvements in quantization performance, particularly for low bits. Additionally, we plan to integrate the KV cache method to address bottleneck issues in scenarios involving long-context.

\section{Limitations}
Despite the advantages, we observed a noticeable gap in accuracy performance for ultra-low bit quantization, particularly with 2-bit quantization, compared to the original model. This challenge could potentially be addressed by exploring non-uniform quantization and mixed-precision quantization, which we leave for future work.

\section{Ethics Statement}
Our research  aims to advance knowledge in LLM quantization. SignRound utilizes open-source models and publicly available datasets, and is not tied to particular applications, requiring only minimal fine-tuning steps on the original models. This ensures that the technical details of our method carry no potential ethical implications. We acknowledge the contributions of the creators and maintainers of these resources and provide citations to the original sources.

\bibliography{custom}

\begin{thebibliography}{64}
\providecommand{\natexlab}[1]{#1}

\bibitem[{Badri and Shaji(2023)}]{badri2023hqq}
Hicham Badri and Appu Shaji. 2023.
\newblock \href {https://mobiusml.github.io/hqq_blog/} {Half-quadratic quantization of large machine learning models}.

\bibitem[{Balles et~al.(2020)Balles, Pedregosa, and Roux}]{balles2020geometry}
Lukas Balles, Fabian Pedregosa, and Nicolas~Le Roux. 2020.
\newblock The geometry of sign gradient descent.
\newblock \emph{arXiv preprint arXiv:2002.08056}.

\bibitem[{Bisk et~al.(2020)Bisk, Zellers, Gao, Choi et~al.}]{bisk2020piqa}
Yonatan Bisk, Rowan Zellers, Jianfeng Gao, Yejin Choi, et~al. 2020.
\newblock Piqa: Reasoning about physical commonsense in natural language.
\newblock In \emph{Proceedings of the AAAI conference on artificial intelligence}, volume~34, pages 7432--7439.

\bibitem[{Bondarenko et~al.(2024)Bondarenko, Nagel, and Blankevoort}]{bondarenko2024quantizable}
Yelysei Bondarenko, Markus Nagel, and Tijmen Blankevoort. 2024.
\newblock Quantizable transformers: Removing outliers by helping attention heads do nothing.
\newblock \emph{Advances in Neural Information Processing Systems}, 36.

\bibitem[{Cheng et~al.(2023)Cheng, Cai, Lv, and Shen}]{cheng2023teq}
Wenhua Cheng, Yiyang Cai, Kaokao Lv, and Haihao Shen. 2023.
\newblock Teq: Trainable equivalent transformation for quantization of llms.
\newblock \emph{arXiv preprint arXiv:2310.10944}.

\bibitem[{Clark et~al.(2019)Clark, Lee, Chang, Kwiatkowski, Collins, and Toutanova}]{clark2019boolq}
Christopher Clark, Kenton Lee, Ming-Wei Chang, Tom Kwiatkowski, Michael Collins, and Kristina Toutanova. 2019.
\newblock Boolq: Exploring the surprising difficulty of natural yes/no questions.
\newblock In \emph{Proceedings of the 2019 Conference of the North American Chapter of the Association for Computational Linguistics: Human Language Technologies, Volume 1 (Long and Short Papers)}, pages 2924--2936.

\bibitem[{Clark et~al.(2018)Clark, Cowhey, Etzioni, Khot, Sabharwal, Schoenick, and Tafjord}]{clark2018think}
Peter Clark, Isaac Cowhey, Oren Etzioni, Tushar Khot, Ashish Sabharwal, Carissa Schoenick, and Oyvind Tafjord. 2018.
\newblock Think you have solved question answering? try arc, the ai2 reasoning challenge.
\newblock \emph{arXiv preprint arXiv:1803.05457}.

\bibitem[{Dagan et~al.(2010)Dagan, Dolan, Magnini, and Roth}]{dagan2010recognizing}
Ido Dagan, Bill Dolan, Bernardo Magnini, and Dan Roth. 2010.
\newblock Recognizing textual entailment: Rational, evaluation and approaches--erratum.
\newblock \emph{Natural Language Engineering}, 16(1):105--105.

\bibitem[{Dettmers et~al.(2022)Dettmers, Lewis, Belkada, and Zettlemoyer}]{dettmers2022gpt3}
Tim Dettmers, Mike Lewis, Younes Belkada, and Luke Zettlemoyer. 2022.
\newblock Gpt3. int8 (): 8-bit matrix multiplication for transformers at scale.
\newblock \emph{Advances in Neural Information Processing Systems}, 35:30318--30332.

\bibitem[{Dettmers et~al.(2023)Dettmers, Pagnoni, Holtzman, and Zettlemoyer}]{dettmers2023qlora}
Tim Dettmers, Artidoro Pagnoni, Ari Holtzman, and Luke Zettlemoyer. 2023.
\newblock Qlora: Efficient finetuning of quantized llms.
\newblock \emph{arXiv preprint arXiv:2305.14314}.

\bibitem[{Esser et~al.(2020)Esser, McKinstry, Bablani, Appuswamy, and Modha}]{esser2020learned}
Steven~K Esser, Jeffrey~L McKinstry, Deepika Bablani, Rathinakumar Appuswamy, and Dharmendra~S Modha. 2020.
\newblock Learned step size quantization.
\newblock In \emph{International Conference on Learning Representations}.

\bibitem[{Frantar and Alistarh(2022)}]{frantar2022optimal}
Elias Frantar and Dan Alistarh. 2022.
\newblock Optimal brain compression: A framework for accurate post-training quantization and pruning.
\newblock \emph{Advances in Neural Information Processing Systems}, 35:4475--4488.

\bibitem[{Frantar et~al.(2022)Frantar, Ashkboos, Hoefler, and Alistarh}]{frantar2022gptq}
Elias Frantar, Saleh Ashkboos, Torsten Hoefler, and Dan Alistarh. 2022.
\newblock Gptq: Accurate post-training quantization for generative pre-trained transformers.
\newblock \emph{arXiv preprint arXiv:2210.17323}.

\bibitem[{Gao et~al.(2020)Gao, Biderman, Black, Golding, Hoppe, Foster, Phang, He, Thite, Nabeshima et~al.}]{gao2020pile}
Leo Gao, Stella Biderman, Sid Black, Laurence Golding, Travis Hoppe, Charles Foster, Jason Phang, Horace He, Anish Thite, Noa Nabeshima, et~al. 2020.
\newblock The pile: An 800gb dataset of diverse text for language modeling.
\newblock \emph{arXiv preprint arXiv:2101.00027}.

\bibitem[{Gao et~al.(2023)Gao, Tow, Abbasi, Biderman, Black, DiPofi, Foster, Golding, Hsu, Le~Noac'h, Li, McDonell, Muennighoff, Ociepa, Phang, Reynolds, Schoelkopf, Skowron, Sutawika, Tang, Thite, Wang, Wang, and Zou}]{eval-harness}
Leo Gao, Jonathan Tow, Baber Abbasi, Stella Biderman, Sid Black, Anthony DiPofi, Charles Foster, Laurence Golding, Jeffrey Hsu, Alain Le~Noac'h, Haonan Li, Kyle McDonell, Niklas Muennighoff, Chris Ociepa, Jason Phang, Laria Reynolds, Hailey Schoelkopf, Aviya Skowron, Lintang Sutawika, Eric Tang, Anish Thite, Ben Wang, Kevin Wang, and Andy Zou. 2023.
\newblock \href {https://doi.org/10.5281/zenodo.10256836} {A framework for few-shot language model evaluation}.

\bibitem[{Gong et~al.(2024)Gong, Liu, Wang, Cai, Zhao, and Yan}]{gong2024makes}
Zhuocheng Gong, Jiahao Liu, Jingang Wang, Xunliang Cai, Dongyan Zhao, and Rui Yan. 2024.
\newblock What makes quantization for large language model hard? an empirical study from the lens of perturbation.
\newblock In \emph{Proceedings of the AAAI Conference on Artificial Intelligence}, volume~38, pages 18082--18089.

\bibitem[{Hassibi et~al.(1993)Hassibi, Stork, and Wolff}]{hassibi1993optimal}
Babak Hassibi, David~G Stork, and Gregory~J Wolff. 1993.
\newblock Optimal brain surgeon and general network pruning.
\newblock In \emph{IEEE international conference on neural networks}, pages 293--299. IEEE.

\bibitem[{Hendrycks et~al.(2020)Hendrycks, Burns, Basart, Zou, Mazeika, Song, and Steinhardt}]{hendrycks2020measuring}
Dan Hendrycks, Collin Burns, Steven Basart, Andy Zou, Mantas Mazeika, Dawn Song, and Jacob Steinhardt. 2020.
\newblock Measuring massive multitask language understanding.
\newblock In \emph{International Conference on Learning Representations}.

\bibitem[{Hu et~al.(2021)Hu, Wallis, Allen-Zhu, Li, Wang, Wang, Chen et~al.}]{hu2021lora}
Edward~J Hu, Phillip Wallis, Zeyuan Allen-Zhu, Yuanzhi Li, Shean Wang, Lu~Wang, Weizhu Chen, et~al. 2021.
\newblock Lora: Low-rank adaptation of large language models.
\newblock In \emph{International Conference on Learning Representations}.

\bibitem[{Jiang et~al.(2023)Jiang, Sablayrolles, Mensch, Bamford, Chaplot, Casas, Bressand, Lengyel, Lample, Saulnier et~al.}]{jiang2023mistral}
Albert~Q Jiang, Alexandre Sablayrolles, Arthur Mensch, Chris Bamford, Devendra~Singh Chaplot, Diego de~las Casas, Florian Bressand, Gianna Lengyel, Guillaume Lample, Lucile Saulnier, et~al. 2023.
\newblock Mistral 7b.
\newblock \emph{arXiv preprint arXiv:2310.06825}.

\bibitem[{Jiang et~al.(2024)Jiang, Sablayrolles, Roux, Mensch, Savary, Bamford, Chaplot, Casas, Hanna, Bressand et~al.}]{jiang2024mixtral}
Albert~Q Jiang, Alexandre Sablayrolles, Antoine Roux, Arthur Mensch, Blanche Savary, Chris Bamford, Devendra~Singh Chaplot, Diego de~las Casas, Emma~Bou Hanna, Florian Bressand, et~al. 2024.
\newblock Mixtral of experts.
\newblock \emph{arXiv preprint arXiv:2401.04088}.

\bibitem[{Kim et~al.(2023{\natexlab{a}})Kim, Hooper, Gholami, Dong, Li, Shen, Mahoney, and Keutzer}]{kim2023squeezellm}
Sehoon Kim, Coleman Hooper, Amir Gholami, Zhen Dong, Xiuyu Li, Sheng Shen, Michael~W Mahoney, and Kurt Keutzer. 2023{\natexlab{a}}.
\newblock Squeezellm: Dense-and-sparse quantization.
\newblock \emph{arXiv preprint arXiv:2306.07629}.

\bibitem[{Kim et~al.(2023{\natexlab{b}})Kim, Henry, Fahim, and Awadalla}]{kim2023finequant}
Young~Jin Kim, Rawn Henry, Raffy Fahim, and Hany~Hassan Awadalla. 2023{\natexlab{b}}.
\newblock Finequant: Unlocking efficiency with fine-grained weight-only quantization for llms.
\newblock \emph{arXiv preprint arXiv:2308.09723}.

\bibitem[{Kingma and Ba(2014)}]{kingma2014adam}
Diederik~P Kingma and Jimmy Ba. 2014.
\newblock Adam: A method for stochastic optimization.
\newblock \emph{arXiv preprint arXiv:1412.6980}.

\bibitem[{Lee et~al.(2023)Lee, Kim, Kwon, and Lee}]{lee2023flexround}
Jung~Hyun Lee, Jeonghoon Kim, Se~Jung Kwon, and Dongsoo Lee. 2023.
\newblock Flexround: Learnable rounding based on element-wise division for post-training quantization.
\newblock \emph{arXiv preprint arXiv:2306.00317}.

\bibitem[{Lee et~al.(2024)Lee, Kim, Yang, Kwon, Yang, Yoo, and Lee}]{lee2024lrq}
Jung~Hyun Lee, Jeonghoon Kim, June~Yong Yang, Se~Jung Kwon, Eunho Yang, Kang~Min Yoo, and Dongsoo Lee. 2024.
\newblock Lrq: Optimizing post-training quantization for large language models by learning low-rank weight-scaling matrices.
\newblock \emph{arXiv preprint arXiv:2407.11534}.

\bibitem[{Lee et~al.(2021)Lee, Yun, Hwang, and Yang}]{lee2021cluster}
Jung~Hyun Lee, Jihun Yun, Sung~Ju Hwang, and Eunho Yang. 2021.
\newblock Cluster-promoting quantization with bit-drop for minimizing network quantization loss.
\newblock In \emph{Proceedings of the IEEE/CVF International Conference on Computer Vision}, pages 5370--5379.

\bibitem[{Li et~al.(2023{\natexlab{a}})Li, Lin, Li, Hong, and Chen}]{li2023faster}
Xiuxian Li, Kuo-Yi Lin, Li~Li, Yiguang Hong, and Jie Chen. 2023{\natexlab{a}}.
\newblock On faster convergence of scaled sign gradient descent.
\newblock \emph{IEEE Transactions on Industrial Informatics}.

\bibitem[{Li et~al.(2023{\natexlab{b}})Li, Bubeck, Eldan, Del~Giorno, Gunasekar, and Lee}]{li2023textbooks}
Yuanzhi Li, S{\'e}bastien Bubeck, Ronen Eldan, Allie Del~Giorno, Suriya Gunasekar, and Yin~Tat Lee. 2023{\natexlab{b}}.
\newblock Textbooks are all you need ii: phi-1.5 technical report.
\newblock \emph{arXiv preprint arXiv:2309.05463}.

\bibitem[{Li et~al.(2021)Li, Gong, Tan, Yang, Hu, Zhang, Yu, Wang, and Gu}]{li2021brecq}
Yuhang Li, Ruihao Gong, Xu~Tan, Yang Yang, Peng Hu, Qi~Zhang, Fengwei Yu, Wei Wang, and Shi Gu. 2021.
\newblock Brecq: Pushing the limit of post-training quantization by block reconstruction.
\newblock \emph{arXiv preprint arXiv:2102.05426}.

\bibitem[{Li et~al.(2022)Li, Guo, Zhu, Zhou, Qiu, Gao, Leng, and Guo}]{li2022efficient}
Zhengyi Li, Cong Guo, Zhanda Zhu, Yangjie Zhou, Yuxian Qiu, Xiaotian Gao, Jingwen Leng, and Minyi Guo. 2022.
\newblock Efficient activation quantization via adaptive rounding border for post-training quantization.
\newblock \emph{arXiv preprint arXiv:2208.11945}.

\bibitem[{Lin et~al.(2023)Lin, Tang, Tang, Yang, Dang, and Han}]{lin2023awq}
Ji~Lin, Jiaming Tang, Haotian Tang, Shang Yang, Xingyu Dang, and Song Han. 2023.
\newblock Awq: Activation-aware weight quantization for llm compression and acceleration.
\newblock \emph{arXiv preprint arXiv:2306.00978}.

\bibitem[{Lin et~al.(2022)Lin, Hilton, and Evans}]{lin2022truthfulqa}
Stephanie Lin, Jacob Hilton, and Owain Evans. 2022.
\newblock Truthfulqa: Measuring how models mimic human falsehoods.
\newblock In \emph{Proceedings of the 60th Annual Meeting of the Association for Computational Linguistics (Volume 1: Long Papers)}, pages 3214--3252.

\bibitem[{Liu et~al.(2023{\natexlab{a}})Liu, Liu, and Cheng}]{liu2023oscillation}
Shih-Yang Liu, Zechun Liu, and Kwang-Ting Cheng. 2023{\natexlab{a}}.
\newblock Oscillation-free quantization for low-bit vision transformers.
\newblock In \emph{International Conference on Machine Learning}, pages 21813--21824. PMLR.

\bibitem[{Liu et~al.(2023{\natexlab{b}})Liu, Oguz, Zhao, Chang, Stock, Mehdad, Shi, Krishnamoorthi, and Chandra}]{liu2023llm}
Zechun Liu, Barlas Oguz, Changsheng Zhao, Ernie Chang, Pierre Stock, Yashar Mehdad, Yangyang Shi, Raghuraman Krishnamoorthi, and Vikas Chandra. 2023{\natexlab{b}}.
\newblock Llm-qat: Data-free quantization aware training for large language models.
\newblock \emph{arXiv preprint arXiv:2305.17888}.

\bibitem[{Liu et~al.(2021)Liu, Wang, Han, Zhang, Ma, and Gao}]{liu2021post}
Zhenhua Liu, Yunhe Wang, Kai Han, Wei Zhang, Siwei Ma, and Wen Gao. 2021.
\newblock Post-training quantization for vision transformer.
\newblock \emph{Advances in Neural Information Processing Systems}, 34:28092--28103.

\bibitem[{Mao et~al.(2024)Mao, Wang, Du, Guan, and Xue}]{mao2024compressibility}
Yu~Mao, Weilan Wang, Hongchao Du, Nan Guan, and Chun~Jason Xue. 2024.
\newblock On the compressibility of quantized large language models.
\newblock \emph{arXiv preprint arXiv:2403.01384}.

\bibitem[{Marcus et~al.(1993)Marcus, Santorini, and Marcinkiewicz}]{marcus1993building}
Mitch Marcus, Beatrice Santorini, and Mary~Ann Marcinkiewicz. 1993.
\newblock Building a large annotated corpus of english: The penn treebank.
\newblock \emph{Computational linguistics}, 19(2):313--330.

\bibitem[{Merity et~al.(2016)Merity, Xiong, Bradbury, and Socher}]{merity2016pointer}
Stephen Merity, Caiming Xiong, James Bradbury, and Richard Socher. 2016.
\newblock Pointer sentinel mixture models.
\newblock In \emph{International Conference on Learning Representations}.

\bibitem[{Mihaylov et~al.(2018)Mihaylov, Clark, Khot, and Sabharwal}]{mihaylov2018can}
Todor Mihaylov, Peter Clark, Tushar Khot, and Ashish Sabharwal. 2018.
\newblock Can a suit of armor conduct electricity? a new dataset for open book question answering.
\newblock In \emph{Proceedings of the 2018 Conference on Empirical Methods in Natural Language Processing}, pages 2381--2391.

\bibitem[{Nagel et~al.(2020)Nagel, Amjad, Van~Baalen, Louizos, and Blankevoort}]{nagel2020up}
Markus Nagel, Rana~Ali Amjad, Mart Van~Baalen, Christos Louizos, and Tijmen Blankevoort. 2020.
\newblock Up or down? adaptive rounding for post-training quantization.
\newblock In \emph{International Conference on Machine Learning}, pages 7197--7206. PMLR.

\bibitem[{Nagel et~al.(2019)Nagel, Baalen, Blankevoort, and Welling}]{nagel2019data}
Markus Nagel, Mart~van Baalen, Tijmen Blankevoort, and Max Welling. 2019.
\newblock Data-free quantization through weight equalization and bias correction.
\newblock In \emph{Proceedings of the IEEE/CVF International Conference on Computer Vision}, pages 1325--1334.

\bibitem[{Paperno et~al.(2016)Paperno, Kruszewski, Lazaridou, Pham, Bernardi, Pezzelle, Baroni, Boleda, and Fern{\'a}ndez}]{paperno2016lambada}
Denis Paperno, Germ{\'a}n Kruszewski, Angeliki Lazaridou, Ngoc-Quan Pham, Raffaella Bernardi, Sandro Pezzelle, Marco Baroni, Gemma Boleda, and Raquel Fern{\'a}ndez. 2016.
\newblock The lambada dataset: Word prediction requiring a broad discourse context.
\newblock In \emph{Proceedings of the 54th Annual Meeting of the Association for Computational Linguistics (Volume 1: Long Papers)}, pages 1525--1534.

\bibitem[{Park et~al.(2022)Park, Park, Kwon, Kim, Lee, and Lee}]{park2022nuqmm}
Gunho Park, Baeseong Park, Se~Jung Kwon, Byeongwook Kim, Youngjoo Lee, and Dongsoo Lee. 2022.
\newblock nuqmm: Quantized matmul for efficient inference of large-scale generative language models.
\newblock \emph{arXiv preprint arXiv:2206.09557}.

\bibitem[{Raffel et~al.(2020)Raffel, Shazeer, Roberts, Lee, Narang, Matena, Zhou, Li, and Liu}]{raffel2020exploring}
Colin Raffel, Noam Shazeer, Adam Roberts, Katherine Lee, Sharan Narang, Michael Matena, Yanqi Zhou, Wei Li, and Peter~J Liu. 2020.
\newblock Exploring the limits of transfer learning with a unified text-to-text transformer.
\newblock \emph{The Journal of Machine Learning Research}, 21(1):5485--5551.

\bibitem[{Safaryan and Richt{\'a}rik(2021)}]{safaryan2021stochastic}
Mher Safaryan and Peter Richt{\'a}rik. 2021.
\newblock Stochastic sign descent methods: New algorithms and better theory.
\newblock In \emph{International Conference on Machine Learning}, pages 9224--9234. PMLR.

\bibitem[{Sakaguchi et~al.(2021)Sakaguchi, Bras, Bhagavatula, and Choi}]{sakaguchi2021winogrande}
Keisuke Sakaguchi, Ronan~Le Bras, Chandra Bhagavatula, and Yejin Choi. 2021.
\newblock Winogrande: An adversarial winograd schema challenge at scale.
\newblock \emph{Communications of the ACM}, 64(9):99--106.

\bibitem[{Shao et~al.(2023)Shao, Chen, Zhang, Xu, Zhao, Li, Zhang, Gao, Qiao, and Luo}]{shao2023omniquant}
Wenqi Shao, Mengzhao Chen, Zhaoyang Zhang, Peng Xu, Lirui Zhao, Zhiqian Li, Kaipeng Zhang, Peng Gao, Yu~Qiao, and Ping Luo. 2023.
\newblock Omniquant: Omnidirectionally calibrated quantization for large language models.
\newblock In \emph{The Twelfth International Conference on Learning Representations}.

\bibitem[{Tang et~al.(2023)Tang, Sun, Wu, Liu, Zhu, and Kang}]{tang2023easyquant}
Hanlin Tang, Yifu Sun, Decheng Wu, Kai Liu, Jianchen Zhu, and Zhanhui Kang. 2023.
\newblock Easyquant: An efficient data-free quantization algorithm for llms.
\newblock In \emph{The 2023 Conference on Empirical Methods in Natural Language Processing}.

\bibitem[{Team et~al.(2024)Team, Mesnard, Hardin, Dadashi, Bhupatiraju, Pathak, Sifre, Rivi{\`e}re, Kale, Love et~al.}]{team2024gemma}
Gemma Team, Thomas Mesnard, Cassidy Hardin, Robert Dadashi, Surya Bhupatiraju, Shreya Pathak, Laurent Sifre, Morgane Rivi{\`e}re, Mihir~Sanjay Kale, Juliette Love, et~al. 2024.
\newblock Gemma: Open models based on gemini research and technology.
\newblock \emph{arXiv preprint arXiv:2403.08295}.

\bibitem[{Touvron et~al.(2023{\natexlab{a}})Touvron, Lavril, Izacard, Martinet, Lachaux, Lacroix, Rozi{\`e}re, Goyal, Hambro, Azhar et~al.}]{touvron2023llama}
Hugo Touvron, Thibaut Lavril, Gautier Izacard, Xavier Martinet, Marie-Anne Lachaux, Timoth{\'e}e Lacroix, Baptiste Rozi{\`e}re, Naman Goyal, Eric Hambro, Faisal Azhar, et~al. 2023{\natexlab{a}}.
\newblock Llama: Open and efficient foundation language models.
\newblock \emph{arXiv preprint arXiv:2302.13971}.

\bibitem[{Touvron et~al.(2023{\natexlab{b}})Touvron, Martin, Stone, Albert, Almahairi, Babaei, Bashlykov, Batra, Bhargava, Bhosale et~al.}]{touvron2023llamav2}
Hugo Touvron, Louis Martin, Kevin Stone, Peter Albert, Amjad Almahairi, Yasmine Babaei, Nikolay Bashlykov, Soumya Batra, Prajjwal Bhargava, Shruti Bhosale, et~al. 2023{\natexlab{b}}.
\newblock Llama 2: Open foundation and fine-tuned chat models.
\newblock \emph{arXiv preprint arXiv:2307.09288}.

\bibitem[{Touvron et~al.(2024)Touvron, Martin, Stone, Albert, Almahairi, Babaei, Bashlykov, Batra, Bhargava, Bhosale et~al.}]{meta2024llama3}
Hugo Touvron, Louis Martin, Kevin Stone, Peter Albert, Amjad Almahairi, Yasmine Babaei, Nikolay Bashlykov, Soumya Batra, Prajjwal Bhargava, Shruti Bhosale, et~al. 2024.
\newblock \href {https://ai.meta.com/blog/meta-llama-3/} {Meta llama 3: The most capable openly available llm to date}.

\bibitem[{Tseng et~al.(2024)Tseng, Chee, Sun, Kuleshov, and De~Sa}]{tseng2024quip}
Albert Tseng, Jerry Chee, Qingyao Sun, Volodymyr Kuleshov, and Christopher De~Sa. 2024.
\newblock Quip\#: Even better llm quantization with hadamard incoherence and lattice codebooks.
\newblock \emph{arXiv preprint arXiv:2402.04396}.

\bibitem[{Wei et~al.(2022)Wei, Gong, Li, Liu, and Yu}]{wei2022qdrop}
Xiuying Wei, Ruihao Gong, Yuhang Li, Xianglong Liu, and Fengwei Yu. 2022.
\newblock \href {https://openreview.net/forum?id=ySQH0oDyp7} {{QD}rop: Randomly dropping quantization for extremely low-bit post-training quantization}.
\newblock In \emph{International Conference on Learning Representations}.

\bibitem[{Wei et~al.(2023)Wei, Zhang, Li, Zhang, Gong, Guo, and Liu}]{Wei_2023}
Xiuying Wei, Yunchen Zhang, Yuhang Li, Xiangguo Zhang, Ruihao Gong, Jinyang Guo, and Xianglong Liu. 2023.
\newblock \href {https://doi.org/10.18653/v1/2023.emnlp-main.102} {Outlier suppression+: Accurate quantization of large language models by equivalent and effective shifting and scaling}.
\newblock In \emph{Proceedings of the 2023 Conference on Empirical Methods in Natural Language Processing}. Association for Computational Linguistics.

\bibitem[{Xiao et~al.(2023)Xiao, Lin, Seznec, Wu, Demouth, and Han}]{xiao2023smoothquant}
Guangxuan Xiao, Ji~Lin, Mickael Seznec, Hao Wu, Julien Demouth, and Song Han. 2023.
\newblock Smoothquant: Accurate and efficient post-training quantization for large language models.
\newblock In \emph{International Conference on Machine Learning}, pages 38087--38099. PMLR.

\bibitem[{Yao et~al.(2021)Yao, Dong, Zheng, Gholami, Yu, Tan, Wang, Huang, Wang, Mahoney et~al.}]{yao2021hawq}
Zhewei Yao, Zhen Dong, Zhangcheng Zheng, Amir Gholami, Jiali Yu, Eric Tan, Leyuan Wang, Qijing Huang, Yida Wang, Michael Mahoney, et~al. 2021.
\newblock Hawq-v3: Dyadic neural network quantization.
\newblock In \emph{International Conference on Machine Learning}, pages 11875--11886. PMLR.

\bibitem[{Yao et~al.(2024)Yao, Wu, Li, Youn, and He}]{yao2024exploring}
Zhewei Yao, Xiaoxia Wu, Cheng Li, Stephen Youn, and Yuxiong He. 2024.
\newblock Exploring post-training quantization in llms from comprehensive study to low rank compensation.
\newblock In \emph{Proceedings of the AAAI Conference on Artificial Intelligence}, volume~38, pages 19377--19385.

\bibitem[{Yuan et~al.(2023)Yuan, Niu, Liu, Liu, Wang, Shang, Sun, Wu, Wu, and Wu}]{yuan2023rptq}
Zhihang Yuan, Lin Niu, Jiawei Liu, Wenyu Liu, Xinggang Wang, Yuzhang Shang, Guangyu Sun, Qiang Wu, Jiaxiang Wu, and Bingzhe Wu. 2023.
\newblock Rptq: Reorder-based post-training quantization for large language models.
\newblock \emph{arXiv preprint arXiv:2304.01089}.

\bibitem[{Yvinec et~al.(2023{\natexlab{a}})Yvinec, Dapogny, and Bailly}]{yvinec2023nupes}
Edouard Yvinec, Arnaud Dapogny, and Kevin Bailly. 2023{\natexlab{a}}.
\newblock Nupes: Non-uniform post-training quantization via power exponent search.
\newblock \emph{arXiv preprint arXiv:2308.05600}.

\bibitem[{Yvinec et~al.(2023{\natexlab{b}})Yvinec, Dapogny, Cord, and Bailly}]{yvinec2023spiq}
Edouard Yvinec, Arnaud Dapogny, Matthieu Cord, and Kevin Bailly. 2023{\natexlab{b}}.
\newblock Spiq: Data-free per-channel static input quantization.
\newblock In \emph{Proceedings of the IEEE/CVF Winter Conference on Applications of Computer Vision}, pages 3869--3878.

\bibitem[{Zellers et~al.(2019)Zellers, Holtzman, Bisk, Farhadi, and Choi}]{Zellers_2019}
Rowan Zellers, Ari Holtzman, Yonatan Bisk, Ali Farhadi, and Yejin Choi. 2019.
\newblock \href {https://doi.org/10.18653/v1/p19-1472} {Hellaswag: Can a machine really finish your sentence?}
\newblock In \emph{Proceedings of the 57th Annual Meeting of the Association for Computational Linguistics}. Association for Computational Linguistics.

\bibitem[{Zhuang et~al.(2021)Zhuang, Tan, Liu, Liu, Reid, and Shen}]{zhuang2021effective}
Bohan Zhuang, Mingkui Tan, Jing Liu, Lingqiao Liu, Ian Reid, and Chunhua Shen. 2021.
\newblock Effective training of convolutional neural networks with low-bitwidth weights and activations.
\newblock \emph{IEEE Transactions on Pattern Analysis and Machine Intelligence}, 44(10):6140--6152.

\end{thebibliography}

% \newpage

\appendix

\section{Quantization Cost}
\label{sec:runtime}

Table \ref{tab:runtime} compares the quantization costs of different methods, with all measurements conducted on a single NVIDIA A100 GPU with 80GB of memory. We ensure each evaluation process exclusively occupies one GPU, but CPU and other resources may be shared among different processes due to limited resources. For SignRound, we disabled low\_gpu\_mem\_usage in our implementation to achieve faster tuning, albeit with higher memory usage. Despite this, LlaMA-2-70B was still able to run on an A100 GPU with 80GB of memory. Although HQQ is exceptionally fast, our methods outperform others in terms of speed. Table \ref{tab:rounding_runtime}  also compares the costs between FlexRound, Adaptive Round, and our method.

\begin{table}[!ht]
\centering
\scalebox{0.76}{
\begin{tabular}{l|lllll}
\hline
\textbf{Model} &  \textbf{GPTQ}  &  \textbf{AWQ}  & \textbf{HQQ} & \textbf{Omni.} & \textbf{Ours} \\  \hline
Llama-2-7B & 1821  &  1328   &  19  & 10255 &  1041  \\ \hline
Llama-2-13B  &  3266   & 2630  &  30  & 18186 &  1918 \\ \hline
Llama-2-70B  &  18517   &  13586  &  119  & 35694  &  9116  \\ \hline
\end{tabular}}
\caption{\label{tab:runtime}
Quantization cost in seconds at W4G-1 for LLaMAV2. Align with the accuracy experiments, OmniQuant 70b is tested with 128 calibration samples, while all the others are tested with 512 samples.}
\end{table}

\section{View of distribution of tuned parameters}

\begin{figure*}[t]
  \centering
  \begin{minipage}{0.45\linewidth}
    \centering
    \includegraphics[width=0.9\linewidth]{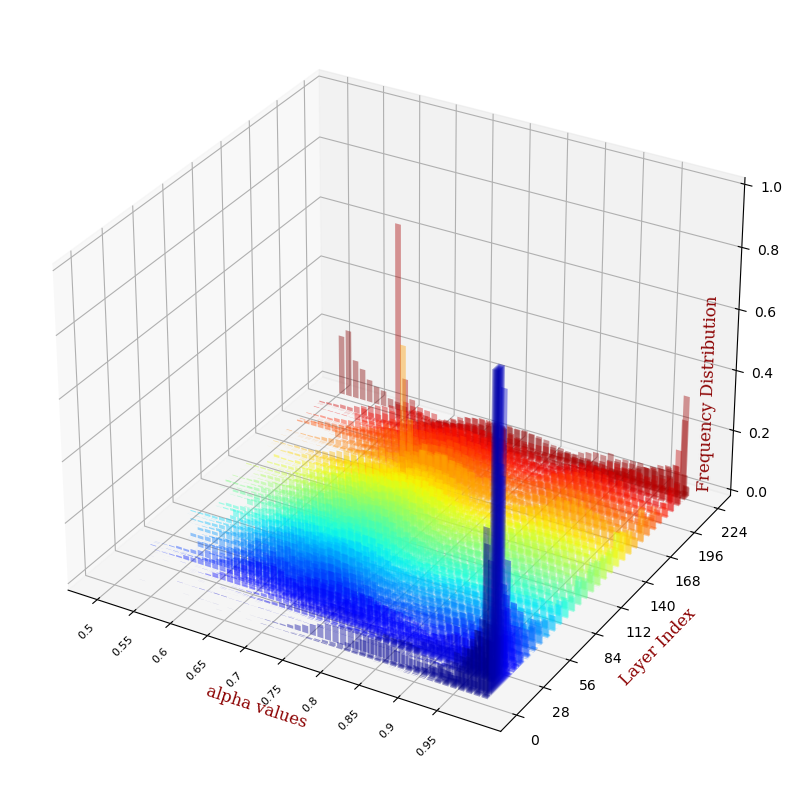}
   \caption*{Mistral-7B, alpha values}
  \end{minipage}
  \begin{minipage}{0.45\linewidth}
    \centering
    \includegraphics[width=0.9\linewidth]{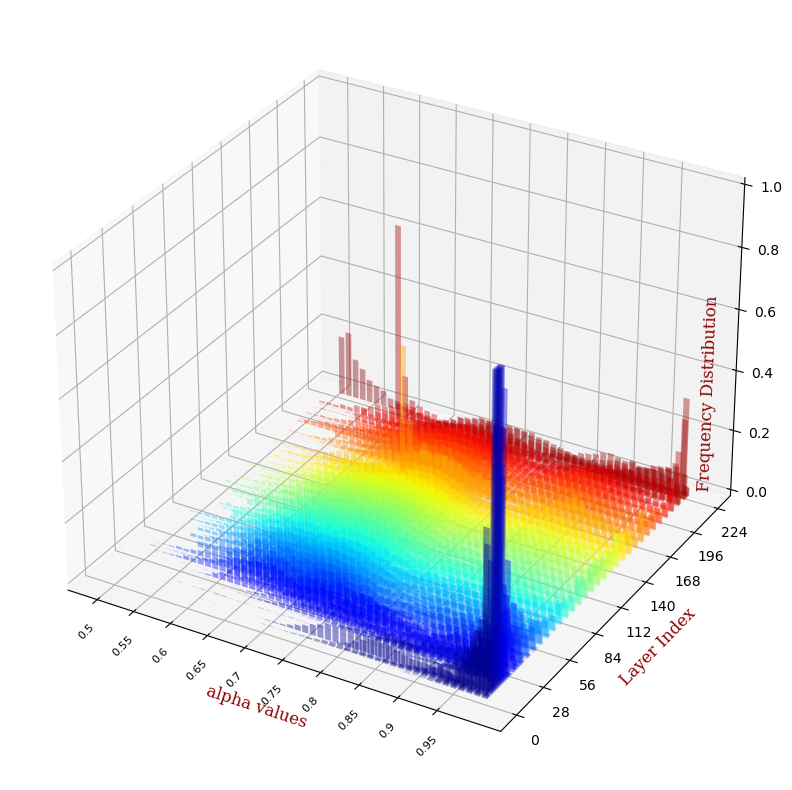}
    \caption*{Llama-2-7B, alpha values}
  \end{minipage}
  \begin{minipage}{0.45\linewidth}
    \centering
    \includegraphics[width=0.9\linewidth]{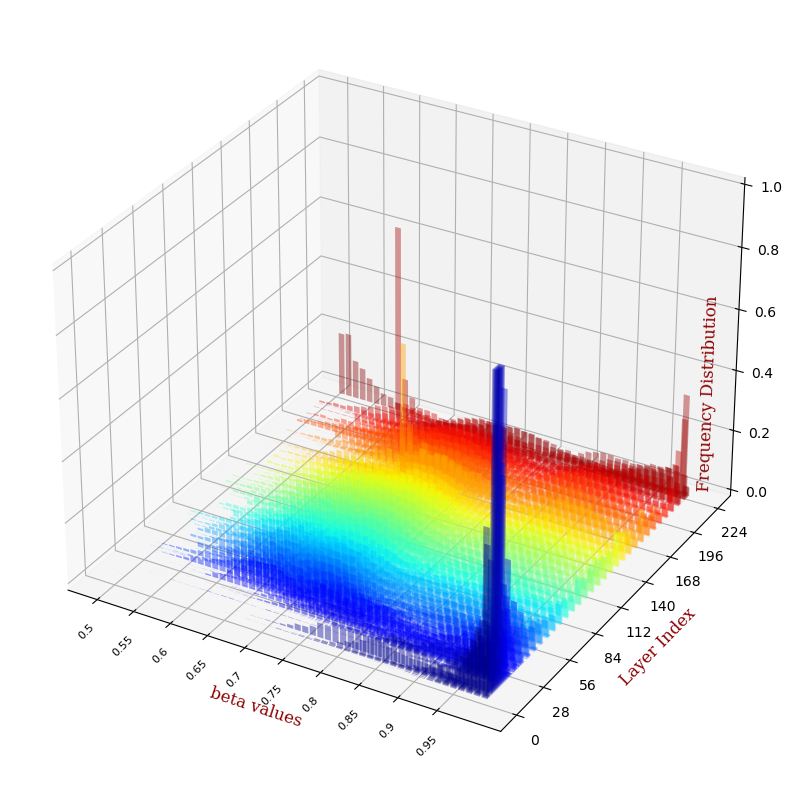}
    \caption*{Mistral-7B, beta values}
  \end{minipage}
  \begin{minipage}{0.45\linewidth}
    \centering
    \includegraphics[width=0.9\linewidth]{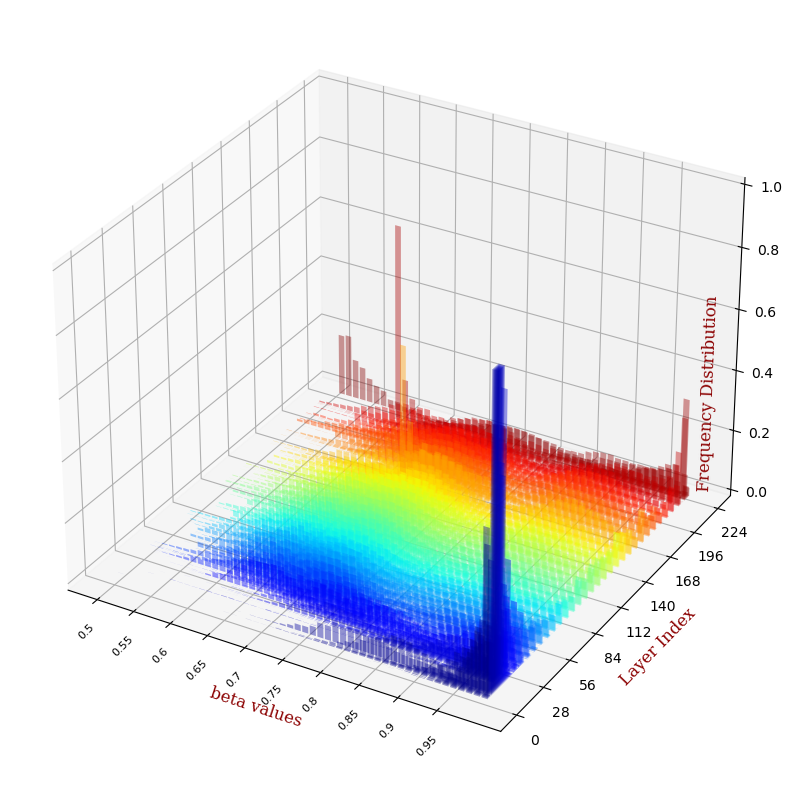}
    \caption*{Llama-2-7B, beta values}
  \end{minipage}
  \begin{minipage}{0.45\linewidth}
    \centering
    \includegraphics[width=0.9\linewidth]{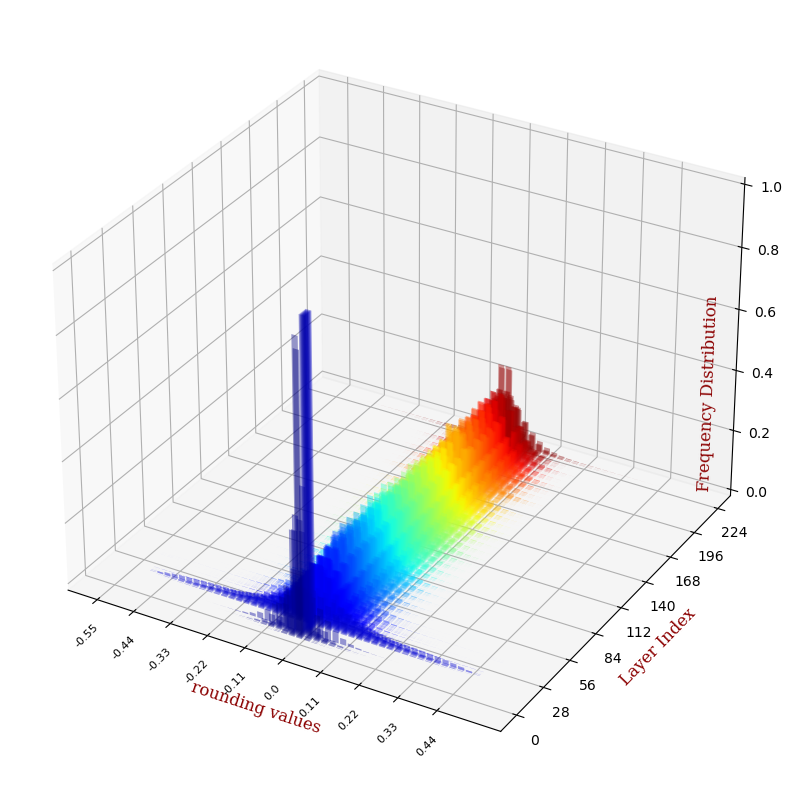}
    \caption*{Mistral-7B,  V values}
  \end{minipage}
  \begin{minipage}{0.45\linewidth}
    \centering
    \includegraphics[width=0.9\linewidth]{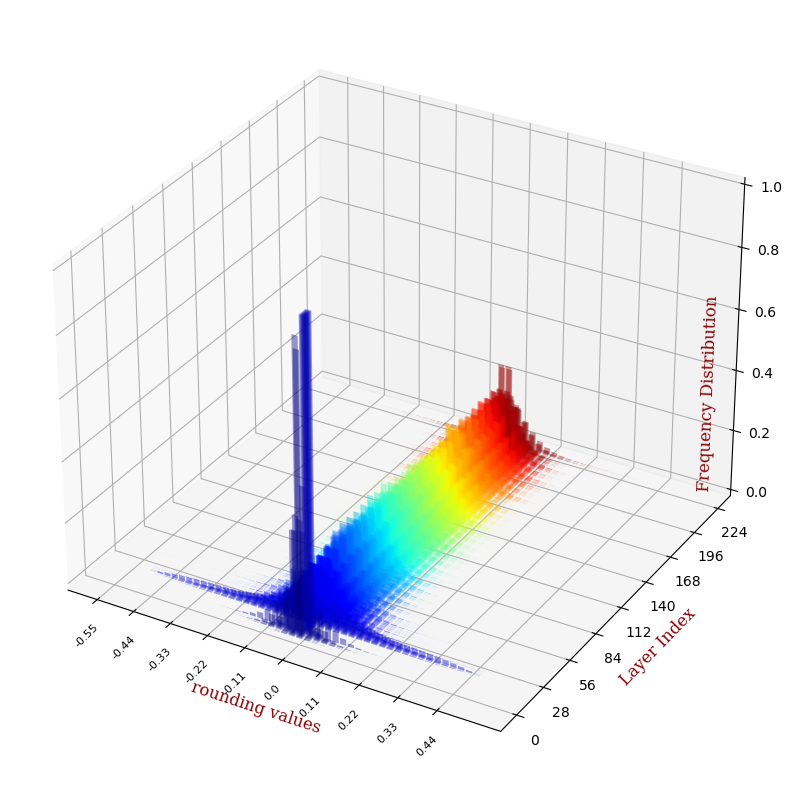}
    \caption*{Llama-2-7B, V values}
  \end{minipage}

\caption{\label{fig:distribution}
The distribution of the magnitude of $\mathbf{V}$ in Eq. \ref{eq 3}  and $\alpha$,  $\beta$ in Eq. \ref{eq 4} for  Mistral-7B-v0.1 and Llama-2-7B at W4G-1, each color in the distribution represents a specific layer index in the models, with blue indicating shallow layers closer to the data layer, and red representing deeper layers.}
\end{figure*}

Figure \ref{fig:distribution} illustrates the distribution of the magnitudes of $\mathbf{V}$ in Eq.\ref{eq 3} and $\alpha$, $\beta$ in Eq. \ref{eq 4}
for Mistral-7B-v0.1 and Llama-2-7B at W4G-1. The results indicate that the distribution is flat for most layers, except for a few layers at the beginning and the end.

\begin{table}[!ht]
\centering
\scalebox{0.76}{
\begin{tabular}{l|lll}
\hline
     \textbf{Model} & \textbf{FlexRound} & \textbf{AdaRound} & \textbf{Ours} \\
     \hline 
    \multirow{1}*{Mistral-7B-V0.1}  &  9369  &  9332  &  1045   \\
    \multirow{1}*{Llama-2-7B}  &  9628  &  9701  &  1041   \\
    \multirow{1}*{Llama-2-13B}  &  17583  &  17865  &  1918   \\
    \hline
\end{tabular}}
\caption{\label{tab:rounding_runtime}
Quantization Time (seconds) of Rounding Methods at W4G-1 with 200 steps for LLaMAV2 Models and Mistral-7B.}
\end{table}

\begin{table*}
\centering
\scalebox{0.86}{
\setlength{\tabcolsep}{1pt}{
\begin{tabular}{ll|cccccccccccccc}
\hline
 \textbf{Model} &  \textbf{Method} &  \textbf{Mmlu} &  \textbf{Lamb.}  &  \textbf{Hella.} &  \textbf{Wino.} &  \textbf{Piqa}  &  \textbf{Truth.} &  \textbf{Open.}  &  \textbf{Boolq}  &  \textbf{RTE} &  \textbf{ARC-e}  &  \textbf{ARC-c.}  &  \textbf{Avg.} \\ \hline
  \multirow{8}*{Mistral-7B} & 16 bits  & 61.35 & 75.68 & 61.27 & 74.03 & 80.79 & 28.03 & 32.80 & 83.67 & 67.51 & 80.81 & 50.34 & 63.30 
    \\
    
    ~ & RTN  & 55.92 & 66.10 & 59.01 & 71.35 & 80.14 & 24.85 & 29.00 & 79.17 & 57.76 & 77.95 & 45.99 & 58.84 
    \\ 
    
    ~ & GPTQ & 58.22 & 73.45 & 59.47 & \textbf{74.03}  & \textbf{80.20}  & 26.93 & 31.00 & 81.50 & 64.98 & 78.24 & 47.01 & 61.37 
      \\ 
       
    ~ & AWQ  & 57.20 & 71.45 & 59.21 & 73.64 & 79.43 & 25.34 & 30.40 & \textbf{82.69}  &\textbf{68.95}  & 79.25 & 47.44 & 61.36 
      \\
    
    ~ &  HQQ & 52.65  & 66.58  & 59.09  & 70.56  & 79.60  & 23.13  & 27.80  & 80.03  & 59.57  & 77.02  & 46.33  & 58.40  \\
    
    ~ &  Omni & 57.52  & 70.00  & 60.27  & 72.93  & 79.87  & 23.99  & 30.80  & 81.53  & 63.90  & 78.54  & 46.42  & 60.52  \\
       
    ~ &  Ours   & \textbf{59.52}  & \textbf{73.76}  & \textbf{60.75}  & 73.32 & 80.09 & \textbf{27.17}  & \textbf{33.00}  & 82.02 & 66.07 & \textbf{80.47}  & \textbf{49.49}  & \textbf{62.33}  \\ 
    
    \cline{3-14}
    
    ~ &  Ours* & \textbf{60.00} & 73.30 & 60.57 & \textbf{74.35} & 80.09 & \textbf{27.91} & 32.20 & \textbf{83.52} & 67.51 & 79.92 & \textbf{49.66} & \textbf{62.64}   \\
    
    \hline
     \multirow{8}*{V2-7B} &   16 bits  &   42.69  &   73.90  &   57.15  &  68.90  &   78.07  &   25.21   &   31.40  &   77.74  &   62.82  &   76.35  &   43.52  &   57.98  \\
    
    ~ &  RTN  &   36.87  &   67.96  &   55.63  &  68.51  &   76.82  &   \textbf{26.19}    &   30.60  &   73.64  &   58.84  &   74.07  &   41.30  &   55.49  \\ 

    ~ & GPTQ  & 39.66 &   \textbf{71.92}  &   55.89  &  68.03  &   77.58    &   25.09  &   30.20  &   76.67  &   62.09  &   75.55  &   41.72  &   56.76  \\
  
    ~ & AWQ &   \textbf{40.24}  &   71.20  &   56.26  &  \textbf{69.61}  &   76.93  &   26.07  &   \textbf{32.60}  &   \textbf{77.31}  &   63.18  &   75.00  &   41.30  &   57.25  \\

    ~ &  HQQ & 28.94  & 43.96  & 48.43  & 59.43  & 71.82  & 23.62  & 24.80  & 52.11  & 53.79  & 64.90  & 34.73  & 46.05  \\

    ~ &  Omni & 39.82  & 71.45  & 55.76  & 67.56  & 76.88  & 25.09  & 30.80  & 76.15  & 64.98  & 74.12  & 40.19  & 56.62  \\
   
    ~ &  Ours  & 39.97 & 71.63 & \textbf{56.52}  & 68.43 & \textbf{77.91}  & 25.70 & 31.60 & 76.18 & \textbf{65.70}  & \textbf{76.01}  & \textbf{42.58}  & \textbf{57.48}  \\

    \cline{3-14}
    
    ~ &  Ours* & \textbf{40.85} & \textbf{72.75} & 56.01 & 67.88 & 77.86 & 25.34 & 31.80 & 76.39 & \textbf{66.43} & 75.88 & 41.55 & \textbf{57.52} \\ \hline
    
    \multirow{8}*{V2-13B} &   16 bits  &   52.86  &   76.77  &   60.04  &  72.14  &   79.05  &   25.95  &   35.20  &   80.55  &   65.34  &   79.38  &   48.38  &   61.42  \\

     ~ & RTN   &   50.37  &   74.35  &   59.12  &  71.98  &   \textbf{79.00}  &   24.85  &   33.00  &   \textbf{81.77}  &   64.98  &   79.08  &   46.59  &   60.46  \\ 
    
     ~ & GPTQ  &   51.14  &   75.37  &   59.14  &  72.06  &   78.02  &   25.34  &   32.20  &   80.46  &   62.09  &   77.36  &   44.54  &   59.79  \\ 
       
     ~ & AWQ  &   51.16  &   \textbf{75.98}  &   59.51  &  70.80  &   78.40  &   25.21  &   \textbf{34.60}  &   78.26  &   \textbf{66.79}  &   79.12  &   46.59  &   60.58  \\
    
      ~ &  HQQ & 35.92  & 49.54  & 46.27  & 58.01  & 72.47  & 23.99  & 19.80  & 61.77  & 51.26  & 62.84  & 33.19  & 46.82  \\
      
    ~ &  Omni & 51.01  & 75.45  & 59.48  & 71.74  & 78.94  & 24.60  & 33.20  & 77.37  & 66.07  & 78.75  & 46.76  & 60.31  \\
       
    ~  & Ours   & \textbf{52.30}  & 75.96 & \textbf{59.79}  & \textbf{72.30}  & 78.84 & \textbf{25.58}  & 34.00 & 80.15 & \textbf{66.79}  & \textbf{79.38}  & \textbf{48.12}  & \textbf{61.20}  \\

    \cline{3-14}
    
    ~ &  Ours* &  52.29 & \textbf{76.15} & 59.73 & 71.90 & 78.51 & 25.21 & 34.40 & 80.24 & \textbf{67.51} & 79.34 & \textbf{48.21} & \textbf{61.23}  \\   \hline
    
     \multirow{7}*{V2-70B} & 16 bits  &   66.23  &   79.64  &   64.77  &  77.98  &   82.15  &   30.60  &   37.20  &   83.70  &   67.87  &   82.70  &   54.44  &   66.12  \\
    
     ~ & RTN   &   63.85  &   77.62  &   63.38  &  76.72  &   81.50  &   28.89  &   \textbf{37.80}  &   83.39  &   68.23  &   81.99  &   54.10  &   65.22  \\
    
    ~ & GPTQ  &   64.81  &   79.27  &   63.86  &  76.87  &   81.61  &   31.46   &   36.40  &   82.23  &   70.04  &   \textbf{82.53}  &   54.18  &   65.75  \\
    
    ~ & AWQ  &   65.08  &   78.77  &   64.14  &  77.11  &   81.45  &   30.48  &   37.20  &   83.64  &   \textbf{72.92}  &   82.49  &   \textbf{55.80}   &   \textbf{66.28} \\
    
    ~ &     HQQ & 56.45  & 66.74  & 53.67  & 73.32  & 76.50  & 25.58  & 33.40  & 67.95  & 61.73  & 72.90  & 43.94  & 57.47  \\
    ~ &     Omni & 64.40  & 79.20  & 63.91  & 76.95  & 81.94  & \textbf{31.70}  & 37.60  & 82.35  & 69.31  & 82.24  & 54.18  & 65.80  \\
       
    ~  & Ours   & \textbf{65.43}  & \textbf{79.55}  & \textbf{64.47}  & \textbf{78.06}  & \textbf{82.10}  & 30.60 & 36.40 & \textbf{83.91}  & 71.12 & \textbf{82.53}  & 54.78 & 66.27  \\ \hline
  \multirow{6}*{V1-7B} & 16 bits    &   32.74  &   73.53  &   56.94  &  70.01  &   78.67  &   22.03  &   34.60  &   75.08  &   66.43  &   75.25  &   41.81  &   57.01  \\

 ~ & RTN  &   31.34  &   70.02  &   55.35  &  69.77  &   77.69  &   20.32  &   32.60  &   73.43  &   59.57  &   74.45  &   41.30  &   55.08  \\ 

 ~ & GPTQ &   29.06  &   71.08  &   55.11  &  70.01  &   77.37  &   20.93  &   32.20  &   72.69  &   63.90  &   74.66  &   41.64  &   55.33  \\
 
~ & AWQ  &   \textbf{33.33}  &   70.81  &   55.98  &  68.27  &   78.07  &   21.18  &   31.40  &   74.37  &   64.62  &   74.03  &   41.21  &   55.75  \\

~  &  Omni  &  32.52  &  \textbf{72.13}  &  55.87  &  \textbf{70.17}  &  78.35  &  \textbf{22.77}  &  32.80  &  75.05  &  66.07  &  \textbf{75.13}  &  40.19  &  56.46  \\  
   
 ~ & Ours  & 31.80 & 71.96  & \textbf{56.57}  & 69.53 & \textbf{79.00}  & 21.91  & \textbf{33.20}  & \textbf{75.72}  & \textbf{66.79}  & 74.83  & \textbf{43.09}  & \textbf{56.76}  \\ \hline
  
\multirow{6}*{V1-13B}  & 16 bits  &   44.21  &   76.21  &   59.92  &  72.77  &   79.16  &   25.70  &   33.20  &   77.89  &   70.76  &   77.40  &   46.42  &   60.33  \\

~ & RTN  &   39.57  &   70.93  &   58.82  &  71.98  &   78.02  &   24.85  &   32.00  &   \textbf{78.20}  &   66.43  &   75.67  &   44.62  &   58.28  \\ 

~  & GPTQ$^+$  &   40.01  &   74.67  &   58.92  &  71.03  &   78.45  &   \textbf{26.44}   &   \textbf{33.60}  &   77.09  &   68.23  &   76.85  &   44.97  &   59.12  \\
   
~ & AWQ  &   \textbf{44.56}  &   74.13  &   59.13  &  71.27  &   \textbf{78.94}  &   25.83  &   33.20  &   76.42  &   66.06  &   76.89  &   \textbf{46.67}   &   59.37   \\

~  &  Omni  &  43.66  &  75.59  &  59.36  &  \textbf{72.38}  &  78.89  &  25.34  &  32.20  &  75.99  &  \textbf{69.68}  &  \textbf{77.10}  &  45.65  &  59.62  \\  
 
~  & Ours  & 43.94 & \textbf{75.82}  & \textbf{59.51}  & 72.22  & 78.78 & 25.70 & 32.80 & 77.34 & 67.51 & 76.47 & \textbf{46.67}  & \textbf{59.71} \\ \hline 

\multirow{6}*{V1-30B}  &   16 bits  &   55.14  &   77.55  &   63.33  &  75.85  &   81.12  &   28.27  &   36.00  &   82.78  &   66.79  &   80.39  &   52.90  &   63.65  \\

~ & RTN   & 53.05 & 75.65 & 62.08 & 74.82 & 80.09 & 25.95 & 35.80 & 81.87 & 63.54 & 79.76 & 50.26 & 62.08  \\ 

~  & GPTQ  &   53.04  &   77.22  &   61.95  &  73.80  &   80.69  &   27.29  &   34.60  &   81.07  &   66.06  &   78.79  &   49.15  &   62.15  \\
   
~ & AWQ  &   54.13  &   76.77  &   62.78  &  74.11  &   \textbf{81.07}  &   \textbf{27.78}   &   35.00  &   \textbf{82.66}  &   67.15  &   79.97  &   51.71  &   63.01  \\

~  &  Omni  &  53.43  &  77.64  &  62.73  &  \textbf{75.30}  &  80.58  &  26.56  &  35.40  &  82.51  &  \textbf{67.87}  &  79.76  &  50.51  &  62.93  \\  
   
~  & Ours & \textbf{54.72}  & \textbf{77.84}  & \textbf{62.91}  & 75.06  & 80.69 & 26.68 & \textbf{36.40}  & 82.60 & 66.79 & \textbf{80.13}  & \textbf{52.13}  & \textbf{63.27}  \\ \hline 

\multirow{6}*{V1-65B} & 16 bits  &   59.79  &   79.12  &   64.53  &  77.35  &   81.23  &   27.91  &   38.00  &   84.86  &   69.68  &   81.36  &   52.82  &   65.15  \\

~ & RTN  &   58.74  &   76.42  &   64.12  &  \textbf{76.72}  &   81.01  &   \textbf{29.25}   &   \textbf{38.60}  &   84.13  &   70.40  &   80.72  &   \textbf{51.88}   &   64.73  \\ 

~ & GPTQ$^+$  &   59.10  &   78.17  &   63.78  &  75.69  &   \textbf{81.34}  &   28.27  &   38.40  &   83.76  &   68.59  &   \textbf{80.98}  &   51.62  &   64.52  \\
   
~ & AWQ  &   58.86  &   77.37  &   63.86  &  76.56  &   80.85  &   28.27  &   35.20  &   83.94  &   \textbf{71.48}  &   78.75  &   50.94  &   64.19  \\

~  &  Omni  &  \textbf{59.59}  &  \textbf{79.16}  &  64.03  &  75.93  &  \textbf{81.99}  &  27.05  &  36.80  &  \textbf{84.65}  &  \textbf{71.48}  &  \textbf{80.98}  &  51.79  &  \textbf{64.86}  \\ 

~  & Ours  & 59.21  & \textbf{79.16}  & \textbf{64.37}  & \textbf{76.64} & 81.34  & 26.81 & 37.80 & 84.40  & 69.68 & \textbf{80.98}  & 51.79 & 64.74  \\ \hline

\end{tabular}}}
\caption{\label{tab:llama and mistral at W4G-1}
Accuracies($\uparrow$) across 11 tasks(0-shot) of LLaMA and Mistral models at W4G-1. The notation GPTQ$^+$ indicates that we adjusted the random seed or data pre-processing to address issues related to the non-positive definite Hessian matrix or other issues.}
\end{table*}

\begin{table*}
\centering
\scalebox{0.9}{
\setlength{\tabcolsep}{1pt}{
\begin{tabular}{ll|cccccccccccccc}
\hline
 \textbf{Model} &  \textbf{Method} &  \textbf{Mmlu} &  \textbf{Lamb.}  &  \textbf{Hella.} &  \textbf{Wino.} &  \textbf{Piqa}  &  \textbf{Truth.} &  \textbf{Open.}  &  \textbf{Boolq}  &  \textbf{RTE} &  \textbf{ARC-e}  &  \textbf{ARC-c.}  &  \textbf{Avg.} \\ \hline
& 16 bits  & 61.35 & 75.68 & 61.27 & 74.03 & 80.79 & 28.03 & 32.80 & 83.67 & 67.51 & 80.81 & 50.34 & 63.30 
 \\
  
& RTN  & 59.72 & 74.44 & \textbf{61.06} & 73.40 & 80.36 & 27.17 & \textbf{32.60} & 83.67 & 64.62 & 79.63 & 49.32 & 62.36 
 \\

& GPTQ & 59.17 & 74.52 & 60.37 & \textbf{74.90} & \textbf{80.58} & 26.68 & 31.00 & 83.33 & \textbf{67.15} & 79.67 & 48.12 & 62.32 
  \\
 
Mistral-7B  & AWQ  & 60.20 & 75.14 & 60.43 & 73.80 & 80.03 & 27.05 & 30.40 & \textbf{84.01} & 62.09 & \textbf{80.39} & \textbf{50.26} & 62.16 
  \\

  & HQQ & 60.02 & 75.41 & 60.79 & 74.11 & 81.01 & 27.29 & 32.60 & 82.97 & 66.79 & 79.92 & 49.32 & 62.75 \\
  & Omni & 59.71 & 73.94 & 60.62 & 73.56 & 80.36 & 26.68 & 30.80 & 83.58 & 65.70 & 80.01 & 49.06 & 62.18 \\
 
 &  Ours  & \textbf{60.47} & \textbf{75.59} & 61.03 & 73.88 & 80.09 & \textbf{27.54} & 31.60 & 83.09 & 66.07 & 79.97 & 49.49 & \textbf{62.62}
  \\ \hline
  &  16 bits  &  42.69 &  73.90 &  57.15 & 68.90 &  78.07 &  25.21  &  31.40 &  77.74 &  62.82 &  76.35 &  43.52  &  57.98  \\
  
& RTN &  40.91 &  72.44 &  \textbf{56.91} & 68.35 &  77.58 &  24.97  &  31.20 &  77.61 &  56.32 &  \textbf{76.26} &  43.52  &  56.92 \\ 

 & GPTQ  &  \textbf{42.57} &  \textbf{73.28} &  56.36 & \textbf{69.06} &  78.02 &  25.34  &  30.20 &  75.72 &  57.04 &  75.63 &  42.15  &  56.85 \\ 
 
  V2-7B & AWQ &  41.00 &  72.60 &  56.40 & 68.98 &  77.31 &  \textbf{25.70}  &  \textbf{31.60} &  \textbf{78.75} &  58.48 &  76.14 &  \textbf{43.86}  &  57.35 \\
  
 & HQQ & 41.79 & 73.20 & 56.21 & 68.43 & 77.58 & 25.83 & 31.60 & 76.09 & 62.82 & 75.84 & 42.15 & 57.41 \\
 
 & Omni & 41.72 & 73.04 & 56.59 & 68.98 & 77.91 & 24.97 & 30.80 & 75.81 & 61.37 & 75.76 & 43.34 & 57.30 \\
 
 &  Ours  & 41.82 & 72.75 & 56.79 & 68.67 & \textbf{78.13} & 25.58 & 30.20 & 77.49 & \textbf{63.54} & 75.76 & 42.58 & \textbf{57.57} \\ \hline

&  16 bits  &  52.86 &  76.77 &  60.04 & 72.14 &  79.05 &  25.95  &  35.20 &  80.55 &  65.34 &  79.38 &  48.38  &  61.42 \\

 &  RTN  &  52.10 &  76.27 &  59.77 & 72.14 &  78.62 &  24.72  &  34.20 &  80.24 &  62.09 &  79.00 &  \textbf{47.95}  &  60.65  \\ 
  & GPTQ  &  \textbf{52.66} &  76.54 &  59.76 & 72.14 &  78.35 &  25.70  &  34.00 &  79.33 &  \textbf{66.43} &  78.58 &  47.53  &  \textbf{61.00} \\ 
 
V2-13B & AWQ  &  52.39 &  \textbf{76.89} &  \textbf{59.97} & \textbf{73.24} &  \textbf{79.00} &  25.21  &  32.60 &  \textbf{80.40} &  63.54 &  79.04 &  47.70  &  60.91 \\

 & HQQ & 52.09 & 75.74 & 59.46 & 72.14 & 78.45 & 24.36 & 33.60 & 79.17 & 66.06 & 79.00 & 47.01 & 60.65 \\
 & Omni & 52.01 & 76.17 & 59.53 & 72.06 & 78.35 & 23.87 & 33.40 & 80.80 & 66.07 & 78.37 & 47.18 & 60.51 \\
    
  & Ours & 51.92 & 76.46 & 59.87 & 71.67 & \textbf{79.00} & \textbf{25.83} & \textbf{35.20} & 79.60 & 63.54 & \textbf{79.25} & 47.01 & 60.85  \\ \hline 

&  16 bits  &  66.23 &  79.64 &  64.77 & 77.98 &  82.15 &  30.60  &  37.20 &  83.70 &  67.87 &  82.70 &  54.44  &  66.12  \\

 &  RTN  &  64.91 &  79.06 &  63.93 & 78.14 &  81.66 &  30.11  &  37.00 &  83.61 &  68.59 &  82.79 &  54.78  &  65.87 \\ 
 & GPTQ &  65.63 &  79.22 &  64.45 & 78.22 &  81.88 &  31.09  &  37.00 &  84.19 &  \textbf{69.31} &  82.79 &  54.61  &  66.22  \\

V2-70B & AWQ &  \textbf{65.79} &  \textbf{79.76} &  64.48 & 77.58 &  \textbf{82.32} &  30.72  &  \textbf{38.00} &  83.06 &  68.95 &  82.70 &  \textbf{55.12}  &  66.23 \\

  & HQQ & 65.34 & 79.14 & 64.56 & 77.35 & 81.56 & 30.48 & 37.20 & 83.67 & 69.31 & 82.83 & 55.20 & 66.06 \\
  
  & Omni & 65.30 & 79.39 & 64.52 & 77.51 & 81.88 & 30.60 & 37.40 & 83.39 & 68.23 & 82.91 & 55.12 & 66.02 \\
 
  & Ours  & 65.65 & 79.49 & \textbf{64.60} & \textbf{78.30} & 82.05 & \textbf{31.58} & 37.40 & \textbf{84.83} & 68.95 & \textbf{82.87} & 54.52 & \textbf{66.39} \\ \hline

 \multirow{6}*{V1-7B} &  16 bits  &  32.74 &  73.53 &  56.94 & 70.01 &  78.67 &  22.03  &  34.60 &  75.08 &  66.43 &  75.25 &  41.81  &  57.01  \\

~ &  RTN  &  32.63 &  72.31 &  56.26 & 70.01 &  78.45 &  20.93  &  \textbf{33.60} &  74.74 &  64.26 &  74.71 &  42.75  &  56.42  \\ 

 ~  & GPTQ  &  31.16 &  72.40 &  55.85 & 70.09 &  78.13 &  \textbf{22.28}  &  30.40 &  74.65 &  64.26 &  74.20 &  40.19  &  55.78  \\
 
~ & AWQ  &  \textbf{33.42} &  \textbf{72.95} &  56.30 & 68.75 &  77.97 &  21.42  &  32.80 &  74.89 &  62.09 &  75.00 &  41.21  &  56.07  \\

 ~  &  Omni  &  31.15  &  72.35  &  56.25  &  69.22  &  78.35  &  21.42  &  33.80  &  74.74  &  65.70  &  74.87  &  \textbf{42.06}  &  56.36  \\  

 ~ & Ours  & 32.15 & 72.85 & \textbf{56.45} & \textbf{70.17} & \textbf{78.51} & \textbf{22.28} & 32.80 & \textbf{75.14} & \textbf{67.87} & \textbf{75.13} & 41.89 & \textbf{56.84}  \\ \hline 
  
  \multirow{6}*{V1-13B} &  16 bits  &  44.21 &  76.21 &  59.92 & 72.77 &  79.16 &  25.70  &  33.20 &  77.89 &  70.76 &  77.40 &  46.42  &  60.33  \\

 ~ &  RTN  &  42.71 &  75.26 &  59.30 & 72.53 &  \textbf{79.54} &  25.95  &  32.60 &  76.76 &  65.34 &  \textbf{76.98} &  45.82  &  59.34 \\ 

 ~  & GPTQ$^+$  &  42.65 &  75.41 &  59.51 & 72.93 &  79.33 &  24.97  &  32.40 &  77.49 &  68.23 &  76.89 &  45.56  &  59.58  \\
 
 ~ & AWQ  &  42.66 &  75.76 &  59.50 & 72.77 &  78.89 &  \textbf{26.56}  &  \textbf{33.60} &  77.46 &  68.59 &  76.94 &  45.48  &  59.84 \\

 ~  &  Omni  &  \textbf{43.99}  &  \textbf{76.29}  &  \textbf{59.53}  &  \textbf{73.56}  &  79.43  &  25.83  &  33.20  &  77.58  &  67.15  &  76.64  &  45.48  &  59.88  \\  

~ & Ours & 42.27 & 76.17 & \textbf{59.53} & \textbf{73.56} & 79.33 & 25.70 & 32.80 & \textbf{78.20} & \textbf{70.04} & 76.94 & \textbf{46.25} & \textbf{60.07}  \\ \hline 

  \multirow{6}*{V1-30B} &  16 bits  &  55.14 &  77.55 &  63.33 & 75.85 &  81.12 &  28.27  &  36.00 &  82.78 &  66.79 &  80.39 &  52.90  &  63.65  \\

 ~ &  RTN  &  54.24 &  77.02 &  62.90 & 74.35 &  \textbf{80.52} &  27.29  &  34.20 &  81.96 &  67.15 &  80.89 &  52.05  &  62.96 \\ 

  ~  & GPTQ  &  54.20 &  77.41 &  62.79 & 75.14 &  80.41 &  27.54  &  34.60 &  81.93 &  67.51 &  80.05 &  50.51  &  62.92 \\
 
~ & AWQ  &  55.14 &  77.49 &  63.08 & \textbf{75.77} &  \textbf{80.52} &  27.29  &  34.20 &  \textbf{82.87} &  67.15 &  80.43 &  \textbf{52.90}  &  63.35  \\

 ~  &  Omni  &  \textbf{55.22}  &  77.80  &  \textbf{63.09}  &  75.14  &  80.30  &  \textbf{28.52}  &  \textbf{36.00}  &  82.20  &  \textbf{69.31}  &  \textbf{80.81}  &  52.82  &  \textbf{63.75}  \\  

 ~ & Ours  & 54.68 & \textbf{77.90} & 62.93 & 74.82 & 80.47 & 28.15 & 35.80 & 82.39 & 66.79 & 80.13 & 51.11 & 63.20  \\ \hline 

 \multirow{6}*{V1-65B} &  16 bits  &  59.79 &  79.12 &  64.53 & 77.35 &  81.23 &  27.91  &  38.00 &  84.86 &  69.68 &  81.36 &  52.82  &  65.15  \\

 ~ &  RTN &  59.53 &  \textbf{79.51} &  64.63 & \textbf{77.35} &  80.96 &  27.91  &  38.40 &  \textbf{84.43} &  \textbf{71.48} &  81.48 &  52.22  &  \textbf{65.26}  \\

  ~  & GPTQ$^+$  &  \textbf{60.47} &  78.79 &  64.45 & 76.24 &  81.18 &  28.03  &  37.40 &  83.85 &  68.95 &  \textbf{81.57} &  \textbf{53.07}  &  64.91  \\
 
~ & AWQ  &  59.45 &  79.31 &  \textbf{64.67} & 76.72 &  \textbf{81.56} &  \textbf{28.15}  &  38.00 &  \textbf{84.43} &  71.12 &  81.10 &  52.13  &  65.15  \\

 ~  &  Omni  &  59.27  &  78.65  &  64.48  &  76.87  &  81.23  &  27.78  &  \textbf{39.00}  &  84.13  &  70.76  &  \textbf{81.57}  &  \textbf{53.07}  &  65.17  \\
 
 ~ & Ours  & 58.93 & 79.22 & 64.48 & 77.03 & 81.28 & 27.91 & 38.60 & 84.31 & 70.76 & 81.19 & 52.22 & 65.08  \\ \hline

\end{tabular}}}
\caption{\label{tab:llama and mistral at W4G128}
Accuracies($\uparrow$) across 11 tasks(0-shot) of LLaMA and Mistral models at W4G128. The notation GPTQ$^+$ indicates that we adjusted the random seed or data pre-processing to address issues related to the non-positive definite Hessian matrix or other issues.}
\end{table*}

\begin{table*}
\centering
\scalebox{0.90}{
\setlength{\tabcolsep}{1pt}{
\begin{tabular}{ll|cccccccccccccc}
\hline
 \textbf{Model} &  \textbf{Method} &  \textbf{Mmlu} &  \textbf{Lamb.}  &  \textbf{Hella.} &  \textbf{Wino.} &  \textbf{Piqa}  &  \textbf{Truth.} &  \textbf{Open.}  &  \textbf{Boolq}  &  \textbf{RTE} &  \textbf{ARC-e}  &  \textbf{ARC-c.}  &  \textbf{Avg.} \\ \hline
 
& 16 bits & 61.35 & 75.68 & 61.27 & 74.03 & 80.79 & 28.03 & 32.80 & 83.67 & 67.51 & 80.81 & 50.34 & 63.30 
  \\
    
& RTN   & 53.49 & 68.74 & 58.12 & 68.27 & 79.33 & 24.60 & 29.60 & 79.97 & 57.40 & 76.89 & 43.77 & 58.20 
  \\ 

& GPTQ  & 55.84 & 73.04 & 57.61 & 70.24 & 78.67 & 24.85 & 30.80 & 81.44 & \textbf{63.54} & 77.27 & 45.65 & 59.91 
 \\ 
   
Mistral-7B  & AWQ  & 55.61 & \textbf{73.69} & 57.86 & 71.27 & \textbf{79.82} & \textbf{26.07} & 29.00 & 81.10 & 59.21 & \textbf{79.00} & \textbf{46.93} & 59.96 
  \\

  & HQQ & 53.97 & 68.66 & 58.59 & 72.22 & 78.73 & 25.70 & 30.00 & 80.24 & 63.90 & 76.81 & 43.86 & 59.33 \\
  
  & Omni & 54.79 & 69.34 & 58.42 & 68.51 & 79.38 & 24.85 & 28.80 & 80.15 & 56.68 & 77.74 & 45.14 & 58.53 \\
   
 &  Ours  & \textbf{57.54} & 73.01 & \textbf{59.60} & \textbf{72.85} & 79.54 & 25.70 & \textbf{31.60} & \textbf{81.74} & 58.12 & 78.70 & 46.33 & \textbf{60.43}
  \\ \hline
  &   16 bits &  42.69 &  73.90 &  57.15 & 68.90 &  78.07 &  25.21  &  31.40 &  77.74 &  62.82 &  76.35 &  43.52  &  57.98  \\
    
&  RTN &  34.22 &  65.96 &  54.90 & 67.56 &  76.28 &  24.48  &  30.80 &  71.68 &  54.51 &  72.98 &  38.57  &  53.81  \\ 

   & GPTQ &  36.11 &  69.61 &  53.66 & \textbf{68.59} &  76.01 &  21.91  &  27.80 &  73.43 &  54.51 &  73.74 &  40.19  &  54.14 \\ 
   
 V2-7B & AWQ &  35.82 &  69.90 &  54.98 & 67.40 &  76.01 &  25.21  &  29.80 &  74.68 &  57.76 &  74.07 &  41.64  &  55.21 \\

 & HQQ & 34.40 & 66.64 & 53.27 & 67.01 & 75.46 & 25.46 & 28.80 & 73.58 & 61.37 & 72.94 & 38.48 & 54.31 \\
 & Omni & 34.51 & 69.75 & 54.42 & 66.69 & 76.77 & 24.24 & 31.40 & 73.21 & 56.68 & 74.37 & 39.85 & 54.72 \\
   
 &  Ours   & \textbf{40.13} & \textbf{71.01} & \textbf{55.33} & 68.27 & \textbf{76.82} & \textbf{25.34} & \textbf{32.80} & \textbf{75.32} & \textbf{60.29} & \textbf{75.25} & \textbf{42.92} & \textbf{56.68} \\ \hline

&   16 bits  &  52.86 &  76.77 &  60.04 & 72.14 &  79.05 &  25.95  &  35.20 &  80.55 &  65.34 &  79.38 &  48.38  &  61.42  \\

 & RTN &  48.01 &  72.33 &  57.74 & 70.72 &  78.07 &  \textbf{25.21}  &  32.00 &  77.28 &  60.65 &  77.69 &  44.62  &  58.57   \\ 
 
   & GPTQ &  49.56 &  \textbf{75.24} &  57.83 & 70.88 &  \textbf{78.56} &  24.97  &  33.40 &  78.44 &  \textbf{62.82} &  77.99 &  45.65  &  \textbf{59.58} \\ 
   
 V2-13B & AWQ  &  \textbf{49.77} &  75.22 &  58.58 & \textbf{71.82} &  77.75 &  24.11  &  \textbf{34.20} &  \textbf{79.97} &  53.43 &  77.95 &  44.62  &  58.86   \\

 & HQQ & 48.40 & 73.22 & 57.66 & 69.77 & 77.31 & 24.11 & 30.60 & 76.97 & 60.29 & 77.15 & 43.60 & 58.10 \\
  & Omni & 47.25 & 73.67 & 58.46 & 70.01 & 78.40 & 24.36 & 33.60 & 79.79 & 64.62 & 77.86 & 46.16 & 59.18 \\
   
  & Ours   & 49.64 & 75.20 & \textbf{59.11} & 71.59 & 78.29 & 24.85 & \textbf{34.20} & 78.47 & 58.12 & \textbf{78.58} & \textbf{45.82} & 59.44  \\ \hline 

&   16 bits &  66.23 &  79.64 &  64.77 & 77.98 &  82.15 &  30.60  &  37.20 &  83.70 &  67.87 &  82.70 &  54.44  &  66.12  \\

 & RTN &  61.15 &  77.95 &  61.98 & \textbf{77.90} &  80.79 &  29.74  &  36.00 &  81.28 &  64.62 &  81.10 &  52.39  &  64.08  \\ 

   & GPTQ &  63.15 &  79.06 &  62.94 & 77.66 &  81.45 &  30.72  &  36.20 &  81.53 &  67.87 &  81.65 &  \textbf{53.67}  &  65.08  \\

 V2-70B & AWQ &  64.09 &  \textbf{79.47} &  63.75 & 76.48 &  \textbf{81.77} &  29.74  &  \textbf{37.20} &  \textbf{82.69} &  66.06 &  81.40 &  \textbf{53.67}  &  65.12 \\

 & HQQ & 63.45 & 78.05 & 63.12 & 77.03 & 81.01 & 29.38 & 36.60 & 82.23 & 66.43 & 81.78 & 53.67 & 64.80 \\
  & Omni & 63.18 & 78.63 & 63.54 & 76.48 & 81.50 & 30.35 & 35.80 & 82.57 & 70.40 & 81.02 & 52.82 & 65.12 \\
  
  & Ours  & \textbf{64.94} & 78.89 & \textbf{63.83} & 76.56 & 81.50 & \textbf{31.21} & \textbf{37.20} & 81.41 & \textbf{68.59} & \textbf{81.73} & 52.56 & \textbf{65.31}   \\ \hline

 \multirow{6}*{V1-7B} &   16 bits  &  32.74 &  73.53 &  56.94 & 70.01 &  78.67 &  22.03  &  34.60 &  75.08 &  66.43 &  75.25 &  41.81  &  57.01  \\

 ~ & RTN  &  28.00 &  67.67 &  53.43 & 66.38 &  76.50 &  21.42  &  31.20 &  72.72 &  59.21 &  70.92 &  38.31  &  53.25  \\ 

   ~ & GPTQ  &  30.16 &  66.31 &  53.92 & 67.48 &  76.82 &  21.42  &  29.60 &  71.31 &  59.21 &  72.22 &  38.74  &  53.38  \\
 
~ & AWQ  &  \textbf{30.33} &  70.19 &  54.53 & 68.98 &  76.71 &  20.81  &  31.60 &  \textbf{74.68} &  64.62 &  73.23 &  38.91  &  \textbf{54.96}  \\

 ~  &  Omni  &  28.35  &  70.54  &  54.48  &  68.27  &  77.48  &  21.05  &  29.40  &  72.29  &  \textbf{66.07}  &  72.73  &  37.12  &  54.34  \\  
   
 ~ & Ours  & 25.85 & \textbf{70.95} & \textbf{55.45} & \textbf{69.69} & \textbf{77.37} & \textbf{21.66} & \textbf{32.00} & 73.88 & 60.29 & \textbf{73.48} & \textbf{39.33} & 54.54  \\ \hline
  
  \multirow{6}*{V1-13B} &   16 bits  &  44.21 &  76.21 &  59.92 & 72.77 &  79.16 &  25.70  &  33.20 &  77.89 &  70.76 &  77.40 &  46.42  &  60.33 \\

  ~  & RTN  &  34.87 &  69.65 &  57.25 & 70.48 &  77.31 &  \textbf{26.93}  &  32.00 &  71.44 &  62.82 &  75.63 &  43.94  &  56.57   \\ 

  ~ & GPTQ  &  35.51 &  73.08 &  57.89 & 70.80 &  77.37 &  24.48  &  31.40 &  \textbf{77.52} &  62.82 &  74.41 &  43.26  &  57.14  \\
   
 ~ & AWQ  &  \textbf{40.53} &  73.94 &  57.89 & 69.53 &  \textbf{78.94} &  26.68  &  33.40 &  74.83 &  \textbf{65.34} &  75.93 &  45.05  &  58.37   \\
 
 ~  &  Omni  &  38.35  &  74.42  &  57.79  &  70.80  &  78.07  &  26.68  &  33.20  &  75.81  &  \textbf{65.34}  &  75.88  &  43.69  &  58.18  \\  

 ~ & Ours  & 39.16 & \textbf{75.22} & \textbf{58.64} & \textbf{71.59} & \textbf{78.94} & 25.95 & \textbf{35.20} & 76.30 & \textbf{65.34} & \textbf{76.52} & \textbf{45.39} & \textbf{58.93}   \\ \hline 

  \multirow{6}*{V1-30B} &   16 bits  &  55.14 &  77.55 &  63.33 & 75.85 &  81.12 &  28.27  &  36.00 &  82.78 &  66.79 &  80.39 &  52.90  &  63.65  \\

~ & RTN  &  52.41 &  75.08 &  61.45 & 74.27 &  79.87 &  25.95  &  33.00 &  81.38 &  65.34 &  79.12 &  48.89  &  61.52 \\ 

 ~  & GPTQ  &  51.39 &  74.97 &  60.35 & \textbf{75.30} &  79.60 &  26.93  &  34.80 &  \textbf{82.75} &  64.62 &  78.11 &  48.46  &  61.57  \\
   
~ & AWQ  &  53.84 &  76.71 &  61.94 & 75.14 &  80.03 &  25.34  &  34.40 &  81.90 &  67.15 &  79.59 &  \textbf{50.77}  &  62.44  \\

   ~  &  Omni  &  53.67  &  76.95  &  61.82  &  74.51  &  80.14  &  25.95  &  34.40  &  81.10  &  66.07  &  79.76  &  48.21  &  62.05  \\  

 ~ & Ours  & \textbf{54.39} & \textbf{77.49} & \textbf{62.13} & 74.03 & \textbf{80.47} & \textbf{27.30} & \textbf{35.00} & \textbf{79.76} & \textbf{68.59} & 79.46 & 48.98 & \textbf{62.51}  \\ \hline 

  \multirow{6}*{V1-65B}  & 16 bits  &  59.79 &  79.12 &  64.53 & 77.35 &  81.23 &  27.91  &  38.00 &  84.86 &  69.68 &  81.36 &  52.82  &  65.15  \\

~ & RTN  &  57.47 &  77.43 &  63.23 & 75.93 &  80.41 &  28.64  &  \textbf{38.40} &  82.69 &  66.43 &  80.22 &  \textbf{51.19}  &  63.82  \\ 

~  & GPTQ$^+$  &  57.92 &  \textbf{78.69} &  62.98 & \textbf{76.87} &  80.63 &  27.66  &  37.60 &  84.16 &  68.95 &  80.89 &  \textbf{51.19}  &  64.32  \\
   
~ & AWQ  &  \textbf{58.87} &  77.94 &  \textbf{63.77} & 75.37 &  \textbf{80.96} &  27.66  &  36.80 &  \textbf{85.02} &  \textbf{71.12} &  \textbf{81.10} &  50.34  &  64.45   \\

~  &  Omni  &  57.19  &  77.00  &  63.15  &  75.53  &  80.90  &  28.15  &  37.60  &  83.18  &  69.68  &  80.18  &  50.51  &  63.92  \\    
   
~  & Ours  & 58.30 & 78.11 & 63.60 & 76.56 & 80.85 & \textbf{29.50} & 37.80 & 84.80 & 70.04 & 80.22 & 50.68 & \textbf{64.59}  \\ \hline
\end{tabular}}}
\caption{\label{tab:llama and mistral at W3G128}
Accuracies($\uparrow$) across 11 tasks(0-shot) of LLaMA and Mistral models at W3G128. The notation GPTQ$^+$ indicates that we adjusted the random seed or data pre-processing to address issues related to the non-positive definite Hessian matrix or other issues.}
\end{table*}

\begin{table*}[ht]
\centering
\scalebox{0.90}{
\setlength{\tabcolsep}{1pt}{
\begin{tabular}{ll|cccccccccccccc}
\hline
 \textbf{Model} &  \textbf{Method} &  \textbf{Mmlu} &  \textbf{Lamb.}  &  \textbf{Hella.} &  \textbf{Wino.} &  \textbf{Piqa}  &  \textbf{Truth.} &  \textbf{Open.}  &  \textbf{Boolq}  &  \textbf{RTE} &  \textbf{ARC-e}  &  \textbf{ARC-c.}  &  \textbf{Avg.} \\ \hline
  
& 16 bits & 61.35 & 75.68 & 61.27 & 74.03 & 80.79 & 28.03 & 32.80 & 83.67 & 67.51 & 80.81 & 50.34 & 63.30 \\
    
& RTN  & 23.45 & 0.14 & 27.43 & 49.64 & 54.30 & 24.24 & 15.20 & 38.69 & 51.99 & 29.08 & 21.59 & 30.52 \\ 

& GPTQ & 25.23 & 30.47 & 38.28 & 53.83 & 64.91 & 24.11 & 17.40 & 58.29 & 50.90 & 47.77 & 24.57 & 39.61 \\ 
   
 Mistral-7B & AWQ  & 25.38 & 0.00 & 25.71 & 52.01 & 51.58 & 23.99 & 17.60 & 37.83 & 47.29 & 26.98 & 22.27 & 30.06 \\

 & HQQ & 23.35 & 0.85 & 27.77 & 51.62 & 56.69 & \textbf{26.68} & 15.80 & 40.55 & 53.43 & 28.62 & 20.14 & 31.41 \\
 
 & Omni & 23.24 & 5.38 & 29.38 & 49.72 & 56.09 & 26.32 & 16.60 & 41.99 & 52.71 & 32.11 & 20.39 & 32.17 \\
   
 &  Ours  & \textbf{40.46} & \textbf{58.61} & \textbf{50.87} & \textbf{62.90} & \textbf{75.84} & 24.85 & \textbf{22.80} & \textbf{78.56} & \textbf{57.04} & \textbf{70.88} & \textbf{37.03} & \textbf{52.71 }
\\ \hline
  &  16 bits  &  42.69 &  73.90 &  57.15 & 68.90 &  78.07 &  25.21  &  31.40 &  77.74 &  62.82 &  76.35 &  43.52  &  57.98  \\
    
&  RTN &  23.98 &  0.02 &  26.04 & 49.49 &  52.50 &  24.85  &  15.20 &  41.01 &  49.10 &  27.48 &  19.71  &  29.94  \\ 

  & GPTQ &  23.65 &  11.72 &  32.59 & 55.17 &  58.32 &  \textbf{25.95}  &  15.80 &  52.14 &  51.99 &  40.45 &  21.25  &  35.37 \\ 
   
 V2-7B & AWQ &  25.38 &  0.00 &  25.69 & 49.96 &  52.34 &  23.75  &  17.80 &  37.83 &  52.71 &  24.62 &  21.08  &  30.10 \\

 & HQQ & 24.51 & 0.02 & 26.06 & 49.49 & 53.26 & 24.72 & 13.80 & 37.92 & 50.90 & 26.52 & 21.33 & 29.87 \\
 
 & Omni & 22.97 & 35.53 & 40.28 & 55.88 & 65.13 & 22.89 & 15.60 & 63.24 & 53.07 & 50.13 & 23.46 & 40.74 \\ 
   
 &  Ours   & \textbf{27.20} & \textbf{55.25} & \textbf{47.35} & \textbf{61.01} & \textbf{72.96} & 24.85 & \textbf{25.60} & \textbf{68.07} & \textbf{54.51} & \textbf{65.99} & \textbf{32.25} & \textbf{48.64} \\ \hline

&   16 bits  &  52.86 &  76.77 &  60.04 & 72.14 &  79.05 &  25.95  &  35.20 &  80.55 &  65.34 &  79.38 &  48.38  &  61.42  \\

 & RTN &  23.77 &  7.47 &  33.08 & 49.01 &  57.94 &  \textbf{26.19}  &  16.00 &  47.74 &  53.43 &  32.03 &  21.93  &  33.51   \\
 
   & GPTQ  &  24.69 &  45.20 &  41.06 & 55.80 &  67.08 &  23.26  &  19.80 &  54.40 &  52.35 &  55.60 &  27.82  &  42.46 \\
   
 V2-13B & AWQ  &  27.04 &  0.00 &  25.80 & 51.85 &  52.99 &  23.62  &  13.60 &  62.17 &  47.29 &  26.22 &  23.12  &  32.16   \\

 & HQQ & 23.48 & 8.17 & 31.27 & 52.17 & 61.86 & 24.85 & 17.20 & 50.46 & 54.51 & 42.85 & 21.25 & 35.28 \\
 
 & Omni & 25.53 & 49.84 & 46.23 & 57.93 & 70.13 & 24.60 & 21.80 & 66.85 & 55.60 & 63.22 & 30.29 & 46.55 \\
   
  & Ours   & \textbf{34.33} & \textbf{63.92} & \textbf{53.35} & \textbf{64.33} & \textbf{76.17} & 25.70 & \textbf{26.00} & \textbf{72.75} & \textbf{61.73} & \textbf{71.17} & \textbf{38.57} & \textbf{53.46}  \\ \hline 

   &   16 bits  & 66.23 & 79.64 & 64.77 & 77.98 & 82.15 & 30.60 & 37.20 & 83.70 & 67.87 & 82.70 & 54.44 & 66.12  \\
    
&  RTN & 24.20 & 20.18 & 40.88 & 54.85 & 63.87 & 24.11 & 17.60 & 43.06 & 53.07 & 50.51 & 27.22 & 38.14  \\ 

   & GPTQ & 23.12 & 0.00 & 25.04 & 49.57 & 49.51 & 0.00 & 27.60 & 37.83 & 52.71 & 25.08 & 22.70 & 28.47 \\ 
   
 V2-70B & AWQ & 24.46 & 0.00 & 25.46 & 51.38 & 52.50 & 23.50 & 14.20 & 62.17 & 52.71 & 25.76 & 22.35 & 32.23 \\

    & HQQ & 23.16 & 19.46 & 35.45 & 56.67 & 66.00 & 22.52 & 20.00 & 40.46 & 52.71 & 52.06 & 23.12 & 37.42 \\
    & Omni & 33.84 & 61.83 & 52.44 & 64.33 & 74.10 & 24.48 & 28.20 & 71.68 & 53.07 & 67.21 & 33.28 & 51.31 \\
   
 &  Ours   & \textbf{54.04} & \textbf{72.97} & \textbf{59.65} & \textbf{74.90} & \textbf{79.00} & \textbf{29.01} & \textbf{34.80 }& \textbf{79.63 }& \textbf{69.68} & \textbf{78.37 }&\textbf{ 46.59} & \textbf{61.69} \\ \hline

 \multirow{6}*{V1-7B} & 16 bits  & 32.74 & 73.53 & 56.94 & 70.01 & 78.67 & 22.03 & 34.60 & 75.08 & 66.43 & 75.25 & 41.81 & 57.01 \\
    
~ & RTN  & 24.36 & 0.52 & 27.24 & 49.25 & 54.24 & 24.24 & 15.20 & 39.63 & \textbf{57.40} & 27.86 & 21.84 & 31.07  \\ 

~ & GPTQ  & 22.95 & 12.75 & 33.36 & 51.70 & 60.07 & 23.99 & 13.40 & 48.62 & 53.07 & 40.82 & 21.50 & 34.75 \\ 
   
~ & AWQ & 23.12 & 0.00 & 25.37 & 53.28 & 52.56 & \textbf{25.21} & 13.80 & 37.83 & 52.71 & 25.63 & 22.53 & 30.18  \\

  ~  &  Omni  &  23.58  &  \textbf{44.23}  &  \textbf{42.39}  &  \textbf{58.48}  &  68.82  &  21.54  &  20.40  &  60.80  &  53.07  &  59.55  &  27.56  &  \textbf{43.68}  \\  

~ &  Ours   & \textbf{24.46} & 13.53 & 42.16 & 56.99 & \textbf{70.02} & 24.60 & \textbf{25.20} & \textbf{62.91} & 47.29 & \textbf{60.90} & \textbf{31.74} & 41.80  \\ \hline

 \multirow{6}*{V1-13B} & 16 bits  & 44.21 & 76.21 & 59.92 & 72.77 & 79.16 & 25.70 & 33.20 & 77.89 & 70.76 & 77.40 & 46.42 & 60.33  \\
    
~ & RTN   & 24.66 & 4.97 & 29.67 & 49.33 & 57.24 & \textbf{25.58} & 12.40 & 44.10 & 53.79 & 32.07 & 22.01 & 32.35  \\ 

~ & GPTQ$^+$  & 26.43 & 40.48 & 39.47 & 58.25 & 66.97 & 23.50 & 18.60 & 52.78 & 50.54 & 51.52 & 25.00 & 41.23  \\ 
   
~ & AWQ  & 27.04 & 0.00 & 25.59 & 50.36 & 53.05 & 24.11 & 15.60 & 62.17 & 47.29 & 25.97 & 23.21 & 32.22  \\

   ~  &  Omni  &  26.93  &  56.41  &  47.67  &  61.17  &  73.23  &  23.38  &  24.60  &  68.75  &  53.07  &  67.00  &  33.79  &  48.73  \\   

~ &  Ours   & \textbf{31.87} & \textbf{59.65} & \textbf{51.25} & \textbf{67.64} & \textbf{76.28} & \textbf{25.58} & \textbf{27.80} & \textbf{69.11} & \textbf{58.48} & \textbf{70.71} & \textbf{37.12} & \textbf{52.32}   \\ \hline

 \multirow{6}*{V1-30B} & 16 bits  & 55.14 & 77.55 & 63.33 & 75.85 & 81.12 & 28.27 & 36.00 & 82.78 & 66.79 & 80.39 & 52.90 & 63.65  \\
    
~ & RTN  & 23.24 & 5.55 & 27.22 & 53.99 & 56.80 & 21.79 & 18.20 & 51.65 & 53.07 & 36.74 & 21.33 & 33.60   \\ 

~ & GPTQ  & 30.47 & 49.93 & 45.05 & 61.88 & 68.88 & 23.26 & 22.60 & 68.29 & 51.99 & 60.69 & 30.72 & 46.70  \\ 
   
~ & AWQ  & 27.04 & 0.00 & 25.41 & 50.20 & 52.94 & \textbf{24.48} & 16.60 & 62.17 & 47.29 & 24.71 & 23.38 & 32.20   \\

~  &  Omni  &  26.89  &  63.03  &  52.23  &  64.64  &  74.27  &  23.87  &  29.20  &  70.86  &  54.51  &  70.45  &  36.18  &  51.47  \\  

~ &  Ours  & \textbf{40.83} & \textbf{67.92} & \textbf{56.73} & \textbf{68.90} & \textbf{76.17} & 24.36 & \textbf{31.60} & \textbf{75.54} & \textbf{62.45} & \textbf{74.92} & \textbf{42.41} & \textbf{56.53}  \\ \hline

 \multirow{6}*{V1-65B} & 16 bits & 59.79 & 79.12 & 64.53 & 77.35 & 81.23 & 27.91 & 38.00 & 84.86 & 69.68 & 81.36 & 52.82 & 65.15  \\
    
~ & RTN  & 24.48 & 32.78 & 43.59 & 57.85 & 67.52 & 22.89 & 22.80 & 61.53 & 50.54 & 52.10 & 28.24 & 42.21   \\ 

~ & GPTQ$^+$  & 37.06 & 67.44 & 53.97 & 69.46 & 76.44 & 24.36 & 28.00 & 73.64 & 60.29 & 71.34 & 38.57 & 54.60 \\ 
   
~ & AWQ  & 25.38 & 0.00 & 25.58 & 49.96 & 53.10 & 24.24 & 11.00 & 37.83 & 52.71 & 24.96 & 22.44 & 29.75  \\

~  &  Omni  &  27.36  &  65.94  &  55.53  &  68.11  &  76.99  &  25.21  &  29.60  &  75.69  &  59.21  &  69.82  &  35.07  &  53.50  \\
   
~ & Ours  & \textbf{47.21} & \textbf{72.07} & \textbf{60.06} &\textbf{73.24} & \textbf{78.62} & \textbf{25.46} & \textbf{34.20} & \textbf{80.64} & \textbf{62.82} & \textbf{77.48} & \textbf{46.76} & \textbf{59.87}  \\ \hline

\end{tabular}}}
\caption{\label{tab:llama and mistral at W2G128}
Accuracies($\uparrow$) across 11 tasks(0-shot) of LLaMA and Mistral models at W2G128. The notation GPTQ$^+$ indicates that we adjusted the random seed or data pre-processing to address issues related to the non-positive definite Hessian matrix or other issues.}
\end{table*}

\begin{table*}
  \centering
    \scalebox{0.68}{
    \begin{tabular}{|c|ll|llll|}
    % \toprule
    \hline
     
    \multicolumn{3}{|c|}{\textbf{LLaMA-V2}} & \textbf{Wiki2.} & \textbf{Ptb}  & \textbf{C4} & \textbf{Wiki.} \\ 
     \cline{1-7}
     
    \multirow{17}*{7B}  & & 16 bits & 5.47 & 37.92 & 7.26 & 8.79 \\
    \cline{2-7}
    ~ & \multirow{5}*{W4G-1} & RTN  & 6.12 & 82.85 & 8.16 & 10.06 \\
                        ~ &  ~ & GPTQ & 5.84 & 1246 & 7.82 & 9.59 \\
                        ~ &  ~ & AWQ  & \textbf{5.81} & \textbf{57.09} & \textbf{7.70} & \textbf{9.42} \\
                        ~ &  ~ & Ours & 7.85 & 3005.52 & 7.71 & 10.34   \\
    \cline{2-7}
    ~ & \multirow{5}*{W4G128} & RTN  & 5.72 & 65.35 & 7.58 & 9.22  \\
                         ~ &   ~ & GPTQ & \textbf{5.60} & 246.28 & 7.48 & 9.05 \\
                          ~ &  ~ & AWQ & 5.61 & \textbf{42.67} & \textbf{7.44} & 9.03   \\ 
                          ~ &  ~ & Ours & 8.96 & 473.78 & 7.50 & \textbf{9.01}  \\ 
    \cline{2-7}
    ~ & \multirow{5}*{W3G128} & RTN  & 6.66 & \textbf{55.10} & 8.98 & 11.21 \\
                         ~ &   ~ & GPTQ & 6.32 & 2245 & 8.55 & 10.37  \\
                          ~ &  ~ & AWQ  & \textbf{6.24} & 66.57 & 8.27 & 10.18  \\
                          ~ &  ~ & Ours & 8.09 & 164.90 & \textbf{8.12} & \textbf{9.76}  \\ 
    \cline{2-7}
    ~ & \multirow{5}*{W2G128} & RTN  & 4270 & 9646 & 4807 & 1.8e5 \\
                         ~ &   ~ & GPTQ & \textbf{25.56} & \textbf{9429} & \textbf{34.87} & \textbf{79.65} \\
                          ~ &  ~ & AWQ  & 2.3e5 & 2.1e5 & 1.7e5 &  1.1e7  \\
                          ~ &  ~ & Ours  & NAN & NAN & NAN & NAN  \\
    \hline
    \multirow{17}*{13B} & & 16 bits & 4.88 & 50.93 & 6.73 & 7.90 \\
    \cline{2-7}
    ~ &\multirow{5}*{W4G-1} & RTN & 5.20 & 60.69 & 7.14 & 8.65 \\
                        ~ &  ~ & GPTQ & 5.12 & 55.99 & 7.04 & 942.3 \\
                        ~ &  ~ & AWQ  & 5.07 & 55.39 & 6.96 & 8.39  \\ 
                        ~ &  ~ & Ours & \textbf{5.00} & \textbf{51.71} & \textbf{6.89} & \textbf{8.33} \\ 
    \cline{2-7}
    ~ & \multirow{5}*{W4G128} & RTN  & 4.98 & 53.69 & 6.87 & 8.12 \\
                         ~ &   ~ & GPTQ & 4.98 & 52.43 & 6.85 & 10.86 \\
                          ~ &  ~ & AWQ & 4.97 & 54.18 & 6.84 & \textbf{8.08} \\ 
                          ~ &  ~ & Ours & \textbf{4.96} & \textbf{51.62} & \textbf{6.83} & 8.14  \\ 
    \cline{2-7}
    ~ & \multirow{5}*{W3G128} & RTN  & 5.52 & 64.85 & 7.58 & 9.27 \\
                         ~ &   ~ & GPTQ & 5.39 & 72.96 & 7.47 & 334.2 \\
                          ~ &  ~ & AWQ  & 5.30 & 57.66 & 7.30 & 8.81   \\ 
                          ~ &  ~ & Ours & \textbf{5.23} & \textbf{53.82} & \textbf{7.18} & \textbf{8.68} \\ 
    \cline{2-7}
    ~ & \multirow{5}*{W2G128} & RTN  & 122.5 & 1212 & 131.8 & 1054 \\
                         ~ &   ~ & GPTQ & 11.30 & \textbf{410.9} & 15.11 & 270.6 \\
                          ~ &  ~ & AWQ  & 1.2e5 & 1.1e5 & 9.7e4 & 5.5e6 \\
                          ~ &  ~ & Ours & \textbf{7.64} & 4250 & \textbf{11.73} & \textbf{57.52}  \\
    \hline
     
    \multirow{17}*{70B}  & & 16 bits & 3.32 & 24.25   & 5.71 & 4.54 \\
    \cline{2-7}
    ~ & \multirow{5}*{W4G-1} & RTN  & 3.67 & \textbf{23.56}   & 6.01 & 5.18 \\
                        ~ &  ~ & GPTQ & 3.57 & 23.76 & 5.89 & 5.00 \\
                        ~ &  ~ & AWQ  & 3.48 & 24.93 & 5.85 & 4.81  \\
                        ~ &  ~ & Ours & \textbf{3.44} & 24.33 & \textbf{5.81} & \textbf{4.78}   \\
    \cline{2-7}
    ~ & \multirow{5}*{W4G128} & RTN  & 3.46 & 24.20 & 5.83 & 4.78  \\
                         ~ &   ~ & GPTQ & 3.42 & 24.01 & 5.78 & 4.71 \\
                          ~ &  ~ & AWQ  & 3.41 & 24.36 & \textbf{5.77} & 4.70   \\ 
                          ~ &  ~ & Ours & \textbf{3.40} & \textbf{23.69} & \textbf{5.77} & \textbf{4.68} \\ 
    \cline{2-7}
    ~ & \multirow{5}*{W3G128} & RTN  & 3.98 & 23.59 & 6.27 & 5.77 \\
                         ~ &   ~ & GPTQ & 3.83 & 24.78 & 6.09 & 5.50  \\
                          ~ &  ~ & AWQ  & 3.73 & 25.68 & 6.03 & 5.31   \\
                          ~ &  ~ & Ours & \textbf{3.68} & \textbf{24.26} & \textbf{5.99} & \textbf{5.23} \\
    \cline{2-7}
    ~ & \multirow{5}*{W2G128} & RTN & \textbf{27.01} & \textbf{758.9} & \textbf{47.57} & \textbf{298.3}   \\
                         ~ &   ~ & GPTQ & NAN & NAN & NAN & NAN \\
                          ~ &  ~ & AWQ & 7.2e4 & 8.1e4 & NAN & 2.5e6 \\
                          ~ &  ~ & Ours & NAN & NAN & NAN & NAN \\ 
    \hline

    \multicolumn{3}{|c|}{\textbf{Mistral}} & \textbf{Wiki2.} & \textbf{Ptb}  & \textbf{C4} & \textbf{Wiki.} \\ 
     \cline{1-7}
      \multirow{17}*{7B}  & & 16 bits & 5.25 & 35.00 & 8.38 & OOM \\
    \cline{2-7}
    ~ &  \multirow{5}*{W4G-1} & RTN  & 5.99 & 44.88 & 9.47 & OOM  \\
                        ~ & ~ & GPTQ & 5.57 & 54.45 & 8.86 & OOM  \\
                        ~ & ~ & AWQ & 5.75 & \textbf{42.21} & 9.14  & OOM \\
                        ~ & ~ & Ours & \textbf{5.43} & 81.67 & \textbf{8.66} & OOM  \\
        \cline{2-7}
    ~ &  \multirow{5}*{W4G128} & RTN  & 5.42 & 34.08 & 8.62  & OOM  \\
                      ~ &  ~ & GPTQ & 5.37 & 37.53 & 8.56  & OOM \\
                      ~ &  ~ & AWQ & 5.37 & 37.12 & 8.55   & OOM \\
                      ~ &  ~ & Ours & \textbf{5.34} & \textbf{36.36} & \textbf{8.51}   & OOM \\
    \cline{2-7}
    ~ &  \multirow{5}*{W3G128} & RTN  & 6.16 & 49.97 & 9.68  & OOM \\
                        ~ & ~ & GPTQ & 5.90 & 49.50 & 9.30   & OOM \\
                        ~ & ~ & AWQ & 5.90 & 51.01 & 9.27  & OOM \\
                        ~ & ~ & Ours & \textbf{5.66} & \textbf{44.50} & \textbf{8.96}  & OOM \\
        \cline{2-7}
    ~ &  \multirow{5}*{W2G128} & RTN  & 1375 & 2351 & 1015 & OOM \\
                        ~ & ~ & GPTQ & 16.59 & 269.2 & 22.38 & OOM  \\
                        ~ & ~ & AWQ & 3.7e4 & 3.4e4 & 3.7e4  & OOM  \\
                        ~ & ~ & Ours & \textbf{8.70} & \textbf{86.08} & \textbf{12.54} & OOM  \\
     \hline
     
    \end{tabular}

    \begin{tabular}{|c|cc|cccc|}
    \hline
    \multicolumn{3}{|c|}{\textbf{LLaMA-V1}} & \textbf{Wiki2.} & \textbf{Ptb}  & \textbf{C4} & \textbf{Wiki.} \\
    \cline{1-7}
     
    \multirow{17}*{7B}  & & 16 bits & 5.68 & 41.15 & 7.34 & 9.49 \\
    \cline{2-7}
    ~ & \multirow{4}*{W4G-1} & RTN & 6.29  & 48.65 & 8.12 & 10.62  \\
                        ~ &  ~ & GPTQ & 6.13 &\textbf{47.18} & 7.93 & 10.32  \\
                        ~ &  ~ & AWQ  & 5.97 & 48.25 & 7.73 & 10.11  \\
                        ~ &  ~ & Ours & \textbf{5.93} & 54.84 & \textbf{7.62} & \textbf{9.91}  \\
    \cline{2-7}
    ~ & \multirow{4}*{W4G128} & RTN  & 5.96 & \textbf{42.33} & 7.70 & 10.00  \\
                         ~ &   ~ & GPTQ & 5.90 & 42.36 & 7.66 & 9.91 \\
                          ~ &  ~ & AWQ & 5.80 & 44.00 & 7.50 & 9.75 \\
                          ~ &  ~ & Ours & \textbf{5.79} & 56.45 & \textbf{7.49} & \textbf{9.74}  \\ 
    \cline{2-7}
    ~ & \multirow{4}*{W3G128} & RTN  & 7.01 & 56.28 & 9.18 & 12.11 \\
                         ~ &   ~ & GPTQ & 6.60 & 53.75 & 8.72 & 11.46  \\
                          ~ &  ~ & AWQ  & 6.32 & 49.27 & 8.21 & 10.81   \\
                          ~ &  ~ & Ours & \textbf{6.28} & \textbf{47.57} & \textbf{8.09} & \textbf{10.55} \\
       \cline{2-7}
    ~ & \multirow{4}*{W2G128} & RTN & 1847 & 6574 & 936.2 & 1.3e4  \\
                         ~ &   ~ & GPTQ & \textbf{28.52} & \textbf{638.3} & \textbf{37.85} & \textbf{128.0}   \\
                          ~ &  ~ & AWQ  & 2.6e5 & 2.8e5 & 2.9e5 & 2.1e7 \\
                          ~ &  ~ & Ours & 641.8 & 824.9 & 2533 & 1876 \\
    % \midrule
    \hline
    
    \multirow{17}*{13B} & & 16 bits & 5.09 & 28.10 & 6.80 & 14.06 \\
    \cline{2-7}
    ~ &\multirow{4}*{W4G-1} & RTN & 5.53 & 29.45 & 7.23 & 37.17 \\
                        ~ &  ~ & GPTQ & 5.34 & 30.23 & 7.09 & 13.09 \\
                        ~ &  ~ & AWQ  & 5.25 & 30.34 & 7.01 & \textbf{12.36}   \\ 
                        ~ &  ~ & Ours & \textbf{5.21} & \textbf{27.81} & \textbf{6.93} & 113.24 \\ 
    \cline{2-7}
    ~ & \multirow{4}*{W4G128} & RTN  & 5.26 & 28.36 & 6.94 & 25.34 \\
                         ~ &   ~ & GPTQ & 5.19 & 29.36 & 6.91 & \textbf{13.33} \\
                          ~ &  ~ & AWQ & 5.19 & 28.34 & 6.90 & 15.25 \\ 
                          ~ &  ~ & Ours & \textbf{5.18} & \textbf{27.80} & \textbf{6.88} & 59.09  \\ 
    \cline{2-7}
    ~ & \multirow{4}*{W3G128} & RTN & 5.88 & 33.10 & 7.86 & 44.06 \\
                         ~ &   ~ & GPTQ & 5.56 & 32.52 & 7.48 & 95.24 \\
                          ~ &  ~ & AWQ  & 5.53 & 29.63 & 7.34 & 22.26   \\ 
                          ~ &  ~ & Ours & \textbf{5.45} & \textbf{28.13} & \textbf{7.21} & \textbf{15.44} \\ 
       \cline{2-7}
    ~ & \multirow{4}*{W2G128} & RTN  & 797.7 & 1695 & 449.1 & 1.5e4  \\
                         ~ &   ~ & GPTQ & 12.13 & 185.8 & NAN & \textbf{546.1} \\
                          ~ &  ~ & AWQ & 2.8e5 & 2.6e5 & 2.4e5 & 1.6e7  \\
                          ~ &  ~ & Ours  & \textbf{8.36} & \textbf{48.93} & \textbf{10.64} & 1773 \\
    \hline
    \multirow{17}*{30B} & & 16 bits & 4.10 & 23.51 & 6.13 & 6.89 \\
    \cline{2-7}
    ~ &\multirow{4}*{W4G-1} & RTN & 4.54 & 25.49 & 6.54 & 8.03 \\
                        ~ &  ~ & GPTQ & 4.41 & 24.22 & 6.40 & 8.50 \\
                        ~ &  ~ & AWQ  & 4.30 & \textbf{24.20} & 6.30 & \textbf{6.88}   \\ 
                        ~ &  ~ & Ours & \textbf{4.23} & 27.97 & \textbf{6.24} & 6.90 \\ 
    \cline{2-7}
    ~ & \multirow{4}*{W4G128} & RTN & 4.23 & \textbf{23.90} & 6.26 & \textbf{7.05} \\
                         ~ &   ~ & GPTQ & 4.24 & 23.92 & 6.23 & 7.73 \\
                          ~ &  ~ & AWQ & 4.22 & 23.98 & 6.21 & 7.29 \\ 
                          ~ &  ~ & Ours & \textbf{4.18} & 31.38 & \textbf{6.20} & 7.39  \\
    \cline{2-7}
    ~ & \multirow{4}*{W3G128} & RTN & 4.87 & 26.99 & 6.85 & NAN \\
                         ~ &   ~ & GPTQ & 4.72 & 25.14 & 6.73 & 8.44 \\
                          ~ &  ~ & AWQ & 4.61 & \textbf{25.05} & 6.56 & \textbf{7.84}   \\
                          ~ &  ~ & Ours & \textbf{4.50} & 67.01 & \textbf{6.47} & 7.90 \\
           \cline{2-7}
    ~ & \multirow{4}*{W2G128} & RTN & 68.40 & 566.8 & 114.2 & 1192   \\
                         ~ &   ~ & GPTQ  & 9.21 & 59.75 & 12.50 & \textbf{21.21}  \\
                          ~ &  ~ & AWQ  & 2.3e5 & 2.2e5 & 2.4e5 &  1.5e7 \\
                          ~ &  ~ & Ours  & \textbf{7.13} & \textbf{55.40} & \textbf{12.02} & 118.7 \\
    \hline
    \multicolumn{3}{|c|}{\textbf{LLaMA-V1}} & \textbf{Wiki2.} & \textbf{Ptb}  & \textbf{C4} & \textbf{Wiki.} \\
    \cline{1-7}
    \multirow{17}*{65B} & & 16 bits & 3.53 & 25.07 & 5.81 & 4.96 \\
    \cline{2-7}
    ~ &\multirow{4}*{W4G-1} & RTN & 3.92 & 28.07 & 6.07 & 5.60 \\
                        ~ &  ~ & GPTQ & 3.79 & 34.82 & 6.00 & 5.46 \\
                        ~ &  ~ & AWQ & 3.72 & 44.83 & 5.96 & 5.30   \\
                        ~ &  ~ & Ours & \textbf{3.65} & \textbf{22.42} & \textbf{5.89} & \textbf{5.19} \\ 
    \cline{2-7}
    ~ & \multirow{4}*{W4G128} & RTN  & 3.67 & 25.61 & 5.90 & 5.21 \\
                         ~ &   ~ & GPTQ & 3.64 & 33.81 & 5.88 & 5.17  \\
                          ~ &  ~ & AWQ & 3.62 & \textbf{24.46} & \textbf{5.87} & 5.14 \\ 
                          ~ &  ~ & Ours & \textbf{3.61} & 35.87 & \textbf{5.87} & \textbf{5.13}  \\
    \cline{2-7}
    ~ & \multirow{4}*{W3G128} & RTN  & 4.25 & 50.00 & 6.33 & 6.25 \\
                         ~ &   ~ & GPTQ & 4.05 & 32.64 & 6.21 & 6.03 \\
                          ~ &  ~ & AWQ & 3.95 & \textbf{23.48} & 6.14 & 5.83  \\ 
                          ~ &  ~ & Ours & \textbf{3.90} & 29.15 & \textbf{6.08} & \textbf{5.69}  \\
           \cline{2-7}
    ~ & \multirow{5}*{W2G128} & RTN  & 15.21 & 276.7 & \textbf{20.03} & 29.39   \\
                         ~ &   ~ & GPTQ  & 6.85 & \textbf{37.79} & NAN & 12.25  \\
                          ~ &  ~ & AWQ  & 7.3e4 & 6.7e4 & 7.4e4  &  NAN  \\
                          ~ &  ~ & Ours & \textbf{5.52} & NAN & NAN  & \textbf{9.25} \\
    \hline
    \end{tabular}}
\caption{\label{tab:llama mistral ppl}
Perplexity(PPL) ($\downarrow$) of  Wikitext2, PTB, C4 and Wikitext tasks for LLaMA and Mistral models. we follow the source code of GPTQ for wikitext2, PTB and C4 PPL evaluation, while for wikitext, we adopt lm-eval-harness \citep{eval-harness}. NAN indicates not a number, while OOM denotes out of memory.}
\end{table*}

\section{More results}
\label{sec:Appendix other results}
We present the detailed LLMs generalization results in Table \ref{tab:details of LLM generalization}, the accuracy is within 1\% of the 16 bit benchmark after simple fine tuning on different types of models. The detailed accuracy results for 11 tasks using the LLaMA and Mistral models, ranging in size from 7B to 70B, at W2-W4 are provided in Tables \ref{tab:llama and mistral at W4G-1}, \ref{tab:llama and mistral at W4G128}, \ref{tab:llama and mistral at W3G128} and \ref{tab:llama and mistral at W2G128}. The detailed perplexity (PPL) results are shown in Table \ref{tab:llama mistral ppl}.  Overall, SignRound demonstrates a clear advantage in accuracy tasks, particularly in ultra-low bit quantization, achieving state-of-the-art performance compared to several popular weight quantization methods. In terms of perplexity (PPL), SignRound outperformed all other methods in 83 out of 124 scenarios, demonstrating its advantages. However, we observed that several quantization algorithms, including SignRound, exhibit sensitivity across different models and tasks. The reason for this sensitivity is detailed in Section \ref{sec:tasks}.

% \newpage
% \vspace{+15cm}
% \begin{center}
% \centering

\begin{table*}
\begin{center}
\scalebox{0.8}{
\setlength{\tabcolsep}{2pt}{
\begin{tabular}{ll|ccccccccccc|c}
\hline
    \textbf{Model} & \textbf{Method} & \textbf{Mmlu} &  \textbf{Lamb.}  & \textbf{Hella.}  &  \textbf{Wino.} &  \textbf{Piqa}  &  \textbf{Truth.} &  \textbf{Open.}  &  \textbf{Boolq}  &  \textbf{RTE} &  \textbf{ARC-e}  &  \textbf{ARC-c.}  &  \textbf{Avg.} \\ \hline
        \multirow{2}*{Gemma-2b}
        & BF16 & 32.87 &  63.44 &  52.73 & 65.04 & 76.71 & 22.03 & 29.80 & 69.27 & 64.26 & 74.20  & 40.19  & 53.69  \\
        ~ & Ours & 32.97 & 63.07 & 51.59 & 65.43 & 76.12 & 22.03 & 30.00 & 69.39 & 63.90 & 73.53  & 39.33 & 53.40  \\ \hline
        \multirow{2}*{Llama-2-7b-chat-hf} 
        & FP16 & 46.40 & 71.05 & 57.80 & 66.38 & 76.39 & 30.23 & 33.40 & 79.76 & 69.68 & 73.82 & 44.20 & 59.01 \\
        ~ & Ours & 45.45 & 70.37 & 57.06 & 66.14 & 76.33 & 30.35 & 32.60 & 80.64 & 72.92 & 73.36 & 43.52 & 58.97 \\ \hline
        \multirow{2}*{Llama-3-8B-Instruct}
        & BF16 & 63.86 & 71.82 & 57.69   & 71.43   & 78.67 & 36.23 & 34.00 & 82.97 & 67.51 & 81.52 & 52.99 & 63.52  \\
        ~ & Ours  & 63.06 & 72.00 & 56.99 & 72.38  & 77.97 & 35.37 & 33.00 & 83.09 & 68.59 & 80.89 & 51.02  & 63.12  \\ \hline
        \multirow{2}*{Mistral-7B-Instruct-v0.2} 
        & BF16  & 59.06 & 71.41 & 66.02 & 73.95 & 80.52 & 52.51 & 36.00 & 85.35 & 70.40 & 81.61 & 54.35 & 66.47 \\
        ~ & Ours & 58.72 & 71.41 & 65.57 & 73.64 & 80.47 & 51.53 & 34.20 & 85.41 & 71.48 & 81.65 & 54.35  & 66.21  \\ \hline
        \multirow{2}*{Mixtral-8x7B} 
        & BF16 & 68.02 & 78.27 & 64.90 & 76.48 & 82.48 & 34.27 & 35.40 & 85.23 & 70.76 & 84.30 & 56.66 & 66.98 \\
        ~ & Ours & 66.93 & 78.25 & 64.59 & 75.14 & 82.10 & 32.19 & 35.60 & 84.74 & 69.31 & 84.30 & 56.48 & 66.33  \\ \hline
        \multirow{2}*{Mixtral-8x7B-Instruct} 
        & BF16 & 68.85 & 77.18 & 67.67 & 76.87 & 83.51 & 49.69 & 36.80 & 88.50 & 71.84 & 86.99 & 62.20 & 70.00 \\
        ~ & Ours & 68.24 & 77.90 & 67.45 & 77.19 & 83.35 & 48.84 & 37.20 & 87.83 & 70.04 & 87.12 & 62.29 & 69.77  \\ \hline
        \multirow{2}*{Phi-3-mini-4k-instruct} 
        & BF16  & 67.97 & 68.08 & 60.64 & 74.03 & 80.30 & 39.53 & 38.80 & 86.21 & 77.98 & 83.54 & 55.72 & 66.62 \\
        ~ & Ours & 66.59 & 67.71 & 59.70 & 74.59 & 79.33 & 37.45 & 38.80 & 85.66 & 79.06 & 82.70 & 56.83 & 66.33   \\ 
        \hline
\end{tabular}}
}
\caption{\label{tab:details of LLM generalization}
 The detail accuracies($\uparrow$) across 11 tasks(detailed in Section \ref{sec:tasks}) with 1000 steps for LLMs at W4G128}
 \end{center}
 \vspace{+20cm}
\end{table*}

\end{document}